\documentclass{article}

\usepackage{graphicx} 
\usepackage{titlesec}
\usepackage[margin=1.0in]{geometry}
\usepackage{float}
\usepackage{booktabs}
\usepackage{multirow}
\usepackage{graphicx}
\usepackage{amsmath}
\usepackage[T1]{fontenc}
\usepackage{lmodern}
\usepackage{microtype}
\usepackage{dsfont}
\usepackage{amssymb}
\usepackage{xcolor}
\usepackage{makecell}
\usepackage{hyperref}
\usepackage{textcomp}
\usepackage{sidecap}
\usepackage{seqsplit}
\usepackage{tabularx}
\usepackage{booktabs}
\usepackage{comment}
\hypersetup{
    colorlinks=true,
    linkcolor=blue,
    urlcolor=blue,
    citecolor=blue
}
\usepackage{siunitx}           
\usepackage[font=small,labelfont=bf]{caption} 
\usepackage{xparse}
\usepackage{xspace}
\usepackage{array}
\usepackage{ragged2e}
\usepackage{authblk} 
\usepackage[numbers,sort&compress]{natbib}

\usepackage[nolist]{acronym}

\titlespacing*{\section}
  {0pt}  
  {2em}  
  {2em}  

\newcommand{\modelname}{Curiosity\xspace}

\providecommand{\ugmprev}{Curiosity-0226-1B}   
\providecommand{\ugmcurr}{Curiosity-0826-4B}   

\newcommand{\eg}{e.g.\xspace}
\newcommand{\ie}{i.e.\xspace}

\title{\ugmcurr: Generative Medical Event Sequence Modeling Across Pediatric and Adult Populations}

\author[1]{Mani Nagarathnam}
\author[1]{Luke Benham}
\author[1]{Paul Blazek}
\author[1]{John Cashy}
\author[1]{Kevin Chung}
\author[1]{John Muenning}
\author[1]{Hank Voeller}
\author[1]{Shane Waxler}
\author[1]{Davis White}
\author[2]{Alison Callahan}
\author[3]{Kevin Dysart}
\author[4]{Andrew Loza}
\author[5]{Michal A. Mankowski}
\author[6]{Mattheus Ramsis}
\author[2]{Nigam H. Shah}
\author[1]{Karen K. Wong}
\author[1]{Rahul Shah}

\date{}

\affil[1]{Epic Systems}
\affil[2]{Department of Medicine, Stanford School of Medicine}
\affil[3]{The Children's Hospital of Philadelphia}
\affil[4]{Department of Biomedical Informatics and Data Science, Yale University}
\affil[5]{Department of Surgery, NYU Grossman School of Medicine}
\affil[6]{Division of Cardiology, Department of Medicine, University of California-San Diego, San Diego, California}

\begin{document}

\maketitle

\begin{acronym}
    \acro{ASCVD}{atherosclerotic cardiovascular disease}
    \acro{ATC}{Anatomical Therapeutic Chemical}
    \acro{BMI}{body mass index}
    \acro{AUROC}{area under the receiver operating characteristic curve}
    \acro{BOS}[\texttt{BOS}]{beginning-of-sequence}
    \acro{CHF}{congestive heart failure}
    \acro{CKD}{chronic kidney disease}
    \acro{Comet}{Cosmos Medical Event Transformer}
    \acro{COPD}{chronic obstructive pulmonary disease}
    \acro{CPT}{Current Procedural Terminology}
    \acro{ECE}{expected calibration error}
    \acro{EHR}{electronic health record}
    \acro{EOS}[\texttt{EOS}]{end-of-sequence}
    \acro{FLOP}{floating point operation}
    \acro{HgbA1c}{hemoglobin A1c}
    \acro{HDL}{high-density lipoprotein}
    \acro{LDL}{low-density lipoprotein}
    \acro{HPB}{hepatopancreatobiliary}
    \acro{ICD-10-CM}{International Classification of Diseases, 10th Revision, Clinical Modification}
    \acro{LLM}{large language model}
    \acro{LOINC}{Logical Observation Identifiers Names and Codes}
    \acro{LOS}{length of stay}
    \acro{MAE}{mean absolute error}
    \acro{MCTD}{mixed connective tissue disease}
    \acro{NASH}{non-alcoholic steatohepatitis}
    \acro{NLP}{natural language processing}
    \acro{PM/DM}{polymyositis/dermatomyositis}
    \acro{PR-AUC}{precision-recall area under the curve}
    \acro{RMSE}{root-mean square error}
    \acro{RMSLE}{root-mean squared log error}
    \acro{RWD}{real-world data}
    \acro{RWE}{real-world evidence}
    \acro{SLE}{systemic lupus erythematosus}
    \acro{T2DM}{type 2 diabetes mellitus}
    \acro{TFLOP}{tera floating point operation}
    \acro{XGBoost}{gradient-boosted decision trees}
\end{acronym}
\begin{abstract}
We introduce \ugmcurr{}, an updated version of the generative medical event foundation model trained on Epic Cosmos that simulates patient timelines to produce zero-shot clinical predictions. \ugmcurr{} incorporates a broader patient population and richer clinical event representations, while generally matching or exceeding \ugmprev{}, the largest model in the previous Curiosity-0226 series \cite{curiosity_v1}, on evaluated tasks supported by both models. Laboratory value prediction improved, with MAE 18\textendash{}46\% lower than that of \ugmprev{}. We evaluated prediction of disease events across pediatric cohorts aged 2\textendash{}17 years. AUROCs for new asthma diagnosis (0.742\textendash{}0.785) and asthma exacerbation (0.829\textendash{}0.852) were comparable to those in the adult reference cohort, while AUROCs for sickle cell crisis (0.820\textendash{}0.847) were lower than in adults (0.915). Explicit representation of documented death enabled new mortality prediction tasks among adults across hospice, inpatient, and ambulatory settings (AUROC 0.821\textendash{}0.942). Scaling-law analysis guided compute-optimal initial pretraining followed by continued pretraining with context extension. The resulting \ugmcurr{} model has more parameters (4.3B versus 1B) and a longer context window (16K versus 8K). It processes 56\% fewer training tokens and requires 2.2 times the training \acp{FLOP} compared with its predecessor. By extending a common zero-shot prediction strategy to pediatric patients and additional outcomes, \ugmcurr{} broadens opportunities to develop clinical and operational applications without training separate task-specific models. Further validation should assess local calibration, performance across patient subgroups and event frequencies, timeline completion rates, and generalization to new settings, alongside prospective evaluation of clinical and operational utility.
\end{abstract}
\acresetall 

\section{Introduction}
\modelname{} is a generative medical event foundation model pretrained on Epic Cosmos\footnote{\url{https://cosmos.epic.com}}, a dataset of de-identified, de-duplicated longitudinal health records for more than 300 million patients~\cite{cosmos_platform,cosmos_rwe}. Since February 2026, the Curiosity-0226 series~\cite{curiosity_v1} of models has been available to researchers at organizations that participate in Cosmos. These models represent a patient's chart as a sequence of discrete medical event tokens and are trained to predict the next event. Sampling the generated event sequences simulates possible continuations of the recorded patient history and enables outcome estimation across a wide range of clinical tasks without task-specific training. Our prior study established that these models improve predictably with model and dataset size and included the largest scaling-law study on real-world patient journeys to date \cite{curiosity_v1}. 

This report documents the second version of \modelname{}. \ugmcurr{}\footnote{Model names include the month and year a model becomes available and its approximate parameter count.} adds mortality prediction and extends clinical prediction and single-encounter generation to pediatric patients. Pediatric evaluations assess asthma onset, asthma exacerbation, sickle cell crisis, hospital length of stay, 30-day readmission, and single-encounter generation. We also assess changes in existing capabilities, including 30-day readmission, length of stay, laboratory value prediction, incident disease, and single-encounter generation, and compare with \ugmprev{}, the largest model of the original Curiosity-0226 series~\cite{curiosity_v1}, where applicable.

To support this broader scope, we expand the training and evaluation population to include 49 million pediatric patients, and we add new data types, including vitals, non-numeric labs, medication attributes, and explicit representations of death and right-censoring. We also refine how existing data is represented and broaden the vocabulary for laboratory tests and procedures.

\ugmcurr{} scales up the model and extends its context window. The updated model has roughly 4.3B parameters and is trained in two phases: compute-optimal pretraining, followed by long-context training with the context window extended from 8{,}192 to 16{,}384 tokens to support the increase in data types and longer time-horizon predictions. Improvements to hyperparameter selection and the training script make pretraining more sample-efficient: compute-optimal training uses roughly 75 tokens per parameter (TPP), versus on the order of 1,000 for the Curiosity-0226 series.
\providecommand{\modelname}{Curiosity}
\providecommand{\cx}{\textasciicircum{}}                  
\providecommand{\tok}[1]{\texttt{\seqsplit{#1}}}          
\providecommand{\ugmprev}{UGM~2025}                      
\providecommand{\ugmcurr}{UGM~2026}                      
\providecommand{\statelabel}[1]{\par\noindent\textbf{#1.}} 
\providecommand{\eg}{e.g.,\ }
\providecommand{\ie}{i.e.,\ }

\begingroup
\emergencystretch=3em  

\section{Tokenization}
\label{sec:tokenization}
The broader capabilities of \ugmcurr{} are supported by adding new event types and revising how some existing events are represented. The changes include those that better represent data from pediatric patients. The complete model vocabulary comprises 14{,}704 tokens across all event, value, temporal, and special-token categories. We describe each of these changes below and illustrate them in Figure \ref{fig:tokenization_changes}. Representations not discussed are unchanged from \ugmprev~\cite{curiosity_v1}.

\subsection{Pediatric patients}
\label{subsec:tok-population}

\modelname{} includes patients of all ages up to 120 years old, including 49M patients younger than 18 years, of whom 14M have their birth encounter documented. Pediatric and neonatal use cases require a tokenization approach that represents the earliest part of the patient's life. The updated tokenization includes births, gestational age, and birth weight, and uses finer age buckets for infants and young children:

\begin{itemize}
  \item \textbf{Birth event.} Because many events can precede a patient's birth in the record, the birth and associated metadata are not placed in the demographic header. Instead, they are inserted into the timeline at the birth instant. A patient with a documented birth always receives the \tok{Patient\cx LifeEvent\cx\cx Birth} token and may receive any, all, or none of the birth-metadata tokens below (birth weight and gestational age).
  \item \textbf{Gestational age.} This is emitted immediately after the birth event token as one token bucketed by completed week (\tok{BirthInfo\cx GestationalAge\cx\cx<week>}), rounding down to the last completed week. Values below 22 weeks and above 44 weeks are collapsed into single edge buckets (\tok{lt22w}, \tok{gt44w}), as such values are rare and frequently reflect data-quality issues.
  \item \textbf{Birth weight.} This is emitted immediately after the birth and gestational age tokens as a component token (\tok{BirthInfo\cx Vital\cx\cx BirthWeight}) followed by a z-score bucket to represent the value, re-using the approach for the numeric laboratory bucket vocabulary (see \autoref{subsub:lab-buckets}).
   \item \textbf{Higher age-token granularity.} For most patients, \emph{age at start of timeline} uses 1-year buckets. For infants and young children, additional sub-year buckets are introduced: one bucket per day for the first week, one per week for the first four weeks, one per four weeks for the remainder of the first year, and one per six months through age five, after which standard 1-year buckets resume. 
  \item \textbf{Separate age and year-at-start tokens.} \emph{Age at start of timeline} and \emph{year at start of timeline} use distinct vocabularies: age is measured in years, with finer pediatric buckets, emitted as \tok{Patient\cx Age\cx\cx 54y}, and year-at-start uses 5-year buckets for better generalization to future calendar dates, emitted as \tok{Patient\cx YearAtStart\cx\cx 2020-2024}.
  \item \textbf{Birth date handling.} Birth date precision varies from an exact instant to the month level, preserving higher precision for neonates and younger infants, and lower precision beyond 6 months of age. Where only the birth month is known, the birth instant is placed conservatively (\eg, the last instant of the known month) so that no future information leaks into earlier positions of the sequence.
\end{itemize}

\subsection{Timeline structure}

\subsubsection{Beginning-of-sequence token position}
\label{subsub:bos}
The \ugmprev{} tokenization scheme placed the \ac{BOS} token after patient demographics, before the first event. For \ugmcurr{}, the \ac{BOS} token is prepended to every patient sequence. 

\subsubsection{Observation end: death and right-censoring}
\label{subsub:observation-end}

In Cosmos, death is recorded as a date rather than a timestamp, and death events can be under-documented in the \ac{EHR} \cite{death_undercount}. The Cosmos data include deaths entered into the EHR; while deaths that occur within the healthcare facility are captured consistently, deaths that occur outside the health system, such as deaths at home, may be missed unless subsequently reported and documented. The date of deaths occurring outside the health system may be less precise as a result of delayed notification and documentation.

\begin{itemize}
  \item \textbf{Known death.} A death token (\tok{Patient\cx LifeEvent\cx\cx RecordedDeath}) is emitted at the last instant of the recorded death date (death is stored as a date only and is pinned to the last instant of that day). Because events may be documented in the EHR with timestamps later than the recorded death, the \ac{EOS} token is placed after the patient's last medical event, excluding time-delta tokens. 
  \item \textbf{Preliminary cause of death.} When a known death carries a linked cause of death diagnosis, that diagnosis is emitted as ordinary diagnosis tokens immediately after the death token and closed with a \tok{Patient\cx LifeEvent\cx\cx EndPreliminaryCauseOfDeath} token. This token disambiguates the cause of death diagnosis from unrelated diagnoses that may fall in the same region of the timeline.
  \item \textbf{Right-censoring / time padding.} Patients with no known death are padded from their last medical event to the observation end instant using six-month time-delta tokens only. A single time delta granularity is used because, in the existing scheme, any time delta other than six months implies that a non-time event follows it. Because only whole six-month tokens are emitted, a patient whose final gap is shorter than six months receives no padding, and no \ac{EOS} token is placed at the end of the pad. Right-censored sequences therefore terminate on a time-delta token when padding is applied, or on the last medical event otherwise; only known-death sequences terminate on \ac{EOS}.
\end{itemize}

Explicit death and right-censoring representations support mortality and survival predictions. The right-censoring padding represents the absence of observed events without specifying whether the patient remained alive or under observation. 
\subsection{Laboratory tests}
\label{subsec:tok-labs}

\subsubsection{Separate laboratory specimen collection and result reporting}

To prevent laboratory values from appearing before their recorded result time, preserve signal for laboratory tests with longer time intervals between specimen collection and result reporting, and preserve information about specimen collection relevant to an encounter, we represent specimen collection and result availability separately:

\begin{itemize}
  \item A \textbf{collection} marker, the laboratory test component token \emph{without} a result value, is emitted at the \emph{specimen collection} instant. 
  \item The laboratory test \textbf{result} (component + value tokens) is moved to the \emph{result} instant.
  \item If a laboratory test has no collection instant, or its collection instant is after its result instant, the collection marker is placed at ($\text{result instant} - 1\,\mathrm{s}$), immediately preceding the result tokens.
\end{itemize}

\subsubsection{Non-numeric laboratory tests}

Non-numeric (categorical) laboratory results are added to the patient timeline and emitted as a component token followed by a result token, mirroring the numeric laboratory result pattern:

\begin{itemize}
  \item \textbf{Component token.} Non-numeric laboratory test components are of the form \tok{NonNumericLab\cx Component\cx\cx<name>}, so numeric and non-numeric labs are separable. The name is based on the \ac{LOINC}\footnote{LOINC\textsuperscript{\textregistered{}} is a registered trademark of Regenstrief Institute, Inc.} Short Name. 
  \item \textbf{Result token.} One token is emitted per distinct non-numeric result value. Values that are masked or null are grouped into a single result token, and result values below a prevalence floor ($<$10{,}000 occurrences in the full dataset) are mapped to an \texttt{Other} result token distinct from the missing-value token.
\end{itemize}

\subsubsection{Laboratory result inclusion threshold}
Laboratory result inclusion uses an event count threshold applied to the laboratory test \emph{name}. The numeric laboratory test threshold of $\geq\text{400{,}000}$ events includes roughly 1{,}000 unique numeric laboratory tests, and the non-numeric laboratory test threshold of $\geq\text{22{,}000}$ events includes roughly 5{,}000 unique non-numeric laboratory tests. Both the numeric and non-numeric thresholds are chosen so as to include $\approx$99.2\% of all laboratory test events.

\subsubsection{Refined percentile buckets for numeric laboratory results}
\label{subsub:lab-buckets}

Numeric laboratory results are discretized into buckets using a shared set of percentile boundaries derived from a standard-normal z-score grid. The numerical bucket boundaries are determined separately for each laboratory result and unit:

\begin{enumerate}
  \item \textbf{Fix the global parameters.} We use 30 buckets: two tail buckets, each covering 0.1\% of the distribution, and 28 interior buckets. The same percentile boundaries are used for every \ac{LOINC} code.
  \item \textbf{Construct the shared percentile boundaries.} On the standard normal distribution, we select lower and upper z-score boundaries corresponding to the 0.1st and 99.9th percentiles. We divide the z-score range between these boundaries into 28 equal-width intervals, then convert all boundaries to percentiles using the standard normal cumulative distribution function. The resulting percentile intervals have at least 0.1\% of the reference distribution's probability mass.
  \item \textbf{Determine laboratory test value boundaries.} For each LOINC--unit pair, we calculate the observed values at the shared percentile levels. These quantiles become the numerical bucket boundaries for that pair. Thus, the percentile levels are shared across labs and units, while the corresponding numerical values differ. Pairs with fewer than 500 observations in the full dataset are excluded.
  \item \textbf{Emit the token.} For each observed result, the laboratory test and unit select the numerical boundaries used to assign the value to a bucket. The emitted token identifies the laboratory test and bucket, but not the unit, so results reported in different units use the same token vocabulary.
\end{enumerate}

\ugmprev{} bucketed numeric laboratory results by decile, giving each laboratory result 10 buckets of equal probability mass. Equal mass concentrates resolution near the center of the distribution; in sparsely observed tails, buckets can span wide value ranges and assign clinically distinct results, such as mildly and critically elevated values, to the same token. This discretization loses within-bucket information when sampling within the bucket to recover a continuous value. Z-score-based buckets increase resolution in the tails without manually selecting quantile boundaries, while imposing a minimum number of observations per bucket to limit sparsity. The same generalized framework is used to construct the birth weight value tokens described above.

\subsection{Vitals}
\label{subsec:tok-vitals}

Selected vitals are sequenced as medical events. Systolic blood pressure, diastolic blood pressure, height, weight, and \ac{BMI} are each emitted as a single bucketed token of the form \tok{Vitals\cx<type>\cx\cx<bucket>}:

\begin{itemize}
  \item \textbf{Sampling.} Vitals are downsampled by vital type to the latest value in each rolling 6-hour window. This prevents vitals from comprising a disproportionate share of the token sequence, particularly for encounter types where measurements occur frequently. 
  \item \textbf{Placement.} Once sampled, vitals tokens are emitted at the recorded instant from the data.
  \item \textbf{Bucketing.} For each vitals data type, bucket edges were based on the observed distribution combined with clinical input, with extra granularity added in ranges relevant to pediatric patients (\autoref{tab:vitals_bins}). BMI uses fine buckets to preserve weight fluctuations within encounters. Buckets are left-edge exclusive, with the highest and lowest buckets capturing values beyond the outermost thresholds. 
  \item \textbf{BMI, height, and weight handling.} Height, weight, and BMI are all tokenized. Weight is necessary for very young children, for whom BMI is generally not considered an appropriate or routinely used measure~\cite{grummerstrawn2010}. BMI is also tokenized when present so that the model can use it directly. BMI is never recomputed when it is not already present in the source.
  \item \textbf{Plausibility filtering.} Out-of-range values are removed using data-quality plausibility bounds (BMI $(0, 204]$ kg/m\textsuperscript{2}; systolic blood pressure $(0, 370]$ mmHg; diastolic blood pressure $(0, 360]$ mmHg; height $(0, 108]$ in; weight $(0, 1400]$ lb). These align with data quality checks used in the broader Cosmos dataset.
\end{itemize}

\subsection{Expanded procedure vocabulary}
\label{subsec:tok-procedures}

Procedures are tokenized from the \ac{CPT}\footnote{CPT\textsuperscript{\textregistered{}} is a registered trademark of the American Medical Association.} code set, using the code received by the contributing health system. Previously, only the 10{,}000 most prevalent procedures were included. For the current model, every procedure with at least 10{,}000 instances in the base dataset is tokenized, which adds clinically important but less frequent procedures such as stem cell transplants.

\subsection{Additional medication attributes}
\label{subsec:tok-medications}

Each medication order event is represented by its drug identity as a set of \ac{ATC} tokens. For example, low-dose aspirin has an ATC code of B01AC06, which is represented by three tokens, \tok{Medication\cx ATC\_1-3\cx\cx B01}, \tok{Medication\cx ATC\_4-5\cx\cx AC}, \tok{Medication\cx ATC\_6-\cx\cx 06}. Three additional tokens capture six medication attributes grouped into three pairs:

\begin{itemize}
  \item \tok{Medication\cx Route\_Form\cx\cx<id>}: administration route paired with the physical form to represent a distinct medication administration modality.
  \item \tok{Medication\cx Dose\_StrengthDoseUnit\cx\cx<id>}: dose amount paired with strength and dose-unit.
  \item \tok{Medication\cx Frequency\_Quantity\cx\cx<id>}: frequency (including optional PRN indicator) paired with total authorized number of dispenses to represent usage and duration.
\end{itemize}

For each ATC code, the set of these unique pairs is collected and IDs are assigned sequentially based on the number of occurrences. For individual elements or pairs with low prevalence (i.e., fewer than 10,000 occurrences in the population), the unique element or pair is considered null in order to maintain a conservative vocabulary set and not train on very low-prevalence events. Most medication attribute pairs were retained, with full fidelity for 99.96\% of route-form pairs, 99.84\% of dose-strength pairs, and 99.77\% of frequency-quantity pairs. Only minor string cleaning was performed to deduplicate similar medication element strings.

Additionally, medications are now identified in the model input as either patient-reported or ordered. For medications with simultaneous documentation instants, reported medications appear first in the sequence, and every block of one or more reported medications is followed by a flag token indicating the preceding medications were patient-reported. Medication administration events are not currently represented. 

\endgroup

\section{Training}
\label{sec:training}

\ugmprev{} established that pretraining loss follows compute-optimal scaling in model size and data, and that downstream task performance improves predictably as loss falls~\cite{curiosity_v1}. For \ugmcurr{} we re-derive these relationships on the new corpus to learn a compute-optimal pretraining budget for a selected model size of approximately 4.3B parameters (\autoref{tab:training_models}). We then continue pretraining it at a longer context than \ugmprev{} to accommodate the richer patient timeline representations with the updated tokenization scheme. 

Training occurs in two stages: initial pretraining at a context length of 8{,}192 tokens, followed by context extension to 16{,}384 tokens and continued pretraining. The initial stage uses 320B tokens, and the subsequent stage uses 435.6B tokens. Thus, \ugmcurr{} is trained on 755.6B tokens for approximately $3.12 \times 10^{22}$ \acp{FLOP} in total. 

\subsection{Compute-optimal initial pretraining}
\label{subsec:train-tokens}

\modelname{} is trained on the tokenized Epic Cosmos corpus described in \autoref{sec:tokenization}. The maximum event time in the training dataset is explicitly set to April 16, 2025. We do not reuse \ugmprev{}'s compute-optimal sizing, as loss is not comparable across different tokenization schemes. Instead we run an updated Chinchilla-style IsoFLOP scaling-laws analysis~\cite{hoffman, msr_scaling} on the \ugmcurr{} corpus, training a range of smaller models across several FLOP budgets to fit how the optimal model size and token count grow with compute. This analysis places the compute-optimal point at about 75 TPP, compared with roughly 1{,}000 TPP estimated in the scaling analysis for the Curiosity-0226 series~\cite{curiosity_v1}.

\begin{figure}[tbp]
    \centering
    \includegraphics[width=\textwidth]{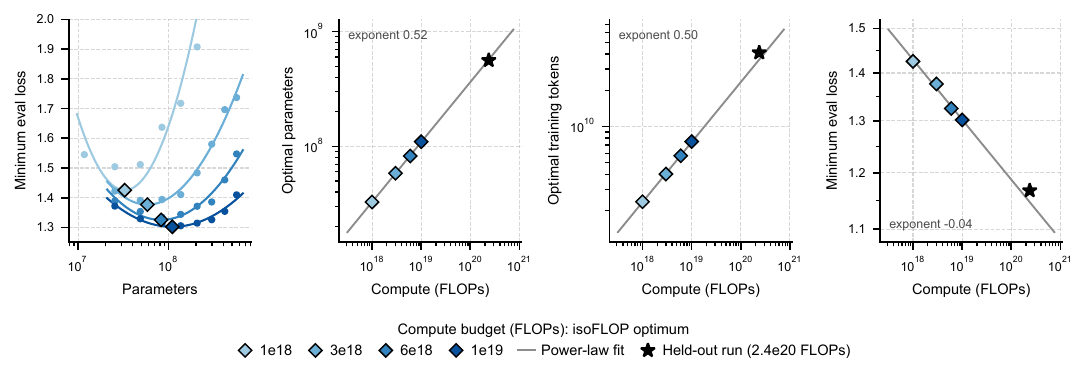}
    \caption[Compute-optimal scaling]{\textbf{Compute-optimal scaling.} Left to right: isoFLOP curves showing minimum evaluation loss against model size for four compute budgets ($10^{18}$ to $10^{19}$ FLOPs), with each run as a point, a quadratic fit in log parameters as a line, and the fitted minimum as a diamond; and the compute-optimal model size, training tokens, and minimum evaluation loss against compute, with power-law fits (gray lines). Optimal model size and training tokens grow at nearly equal rates with compute (exponents 0.52 and 0.50), corresponding to roughly 70--80 training tokens per parameter at the optimum. The star marks a held-out 562M-parameter run at $2.4 \times 10^{20}$ FLOPs that was not used in the fits. In the leftmost panel, three runs at $10^{18}$ FLOPs with loss above 2.0 lie outside the plotted range but are included in the fit.}
    \label{fig:scaling_laws}
\end{figure}

We select the initial pretraining configuration in two steps. First, we relate downstream task performance to pretraining loss: across these scaling-laws models, performance on held-out clinical tasks follows a predictable, roughly sigmoidal function of evaluation loss. Fitting this relationship indicates that an evaluation loss near $1.0$ corresponds to performance that matches or exceeds that of the largest model in the Curiosity-0226 series (\ugmprev{}). Second, we translate that target loss into a compute budget from the fitted scaling laws and, holding the TPP fixed, search over model depth for the configuration that reaches it. The result is a model with approximately 4.3B parameters and 40 transformer layers. Initial pretraining processes 320B tokens, or about one epoch of the corpus. This compute-optimal pretraining uses about $1.14 \times 10^{22}$ FLOPs.

The training recipe for the new corpus includes sweeping key hyperparameters, such as learning rate, batch size, and the model's aspect ratio (the ratio of its hidden width to its number of layers, which trades width against depth at a fixed parameter count). We also replace the previous sequential data loader with one that shuffles across shards to decorrelate training batches. Each of these additions measurably lowers evaluation loss at fixed compute-optimal model sizes.

\subsection{Continued pretraining with context extension}
\label{subsec:train-context-extension}

The initial model is pretrained at the \ugmprev{} context length of 8{,}192 tokens. That window is sufficient for short-horizon tasks, but longer-horizon predictions are frequently truncated, especially as the average number of tokens per patient increases with the new tokenized data types. Much of a patient's history does not fit the previous context size, leaving little room to generate. \citet{wornow} found that longer context improved clinical prediction performance for some EHR foundation model architectures. \ugmcurr{} extends the context window to 16{,}384 tokens.

Extending context on the 4B model required changing how we handled the rotary position embeddings (RoPE)~\cite{RoFormer}. YaRN~\cite{peng2024yarn}, which worked for smaller models, induced catastrophic forgetting of encounter structure at 4B, causing the model to omit encounter-end tokens as prompts grew long, and reduced performance on tasks such as length-of-stay prediction. The final model instead uses the original, unscaled RoPE frequencies, so every position within the original 8{,}192-token window is encoded exactly as in pretraining. We found this better preserved encounter structure at long context. The model then continues pretraining at the new window for 1{,}000 steps followed by 1.5 epochs over the dataset, processing an additional 435.6B tokens and using about $1.98 \times 10^{22}$ FLOPs.

\autoref{tab:training_models} compares training budgets for \ugmcurr{} and \ugmprev{}. Although we re-estimated \ugmcurr{}'s model size and token allocation for the new corpus, initial pretraining uses compute similar to that of \ugmprev{}. In total (initial and continued training), \ugmcurr{} uses approximately 755.6B tokens for about $3.12 \times 10^{22}$ FLOPs in  compute.

\begin{table}[tbp]
\centering
\begin{tabular}{@{}lrrrr@{}}
\toprule
Model / training stage
& Parameters
& \shortstack{Context length\\(tokens)}
& \shortstack{Training\\tokens}
& \shortstack{Compute\\(FLOPs)} \\
\midrule
\textbf{\ugmprev{}}
& 1B & 8{,}192 & 1.7T & $1.4 \times 10^{22}$ \\
\addlinespace
\textbf{\ugmcurr{} (total)}
& 4.3B &  & 755.6B & $3.12 \times 10^{22}$ \\
\quad Initial pretraining
& & 8{,}192 & 320B & $1.14 \times 10^{22}$ \\
\quad Continued pretraining
& & 16{,}384 & 435.6B & $1.98 \times 10^{22}$ \\
\bottomrule
\end{tabular}
\caption{\textbf{Training budgets.} Training budgets for \ugmcurr{} and \ugmprev{}, the largest model of the Curiosity-0226 series~\cite{curiosity_v1}.}
\label{tab:training_models}
\end{table}

\section{Evaluations}
\label{sec:evaluation}

\ugmcurr{} extends prediction capabilities to mortality and pediatric patients (\autoref{subsec:eval-new}), while generally maintaining or improving predictive performance relative to \ugmprev{} on shared evaluation tasks (\autoref{subsec:eval-existing}). 

\subsection{Performance on new capabilities}
\label{subsec:eval-new}

\subsubsection{Mortality prediction}
\label{subsubsec:eval-mortality}

We evaluate the ability of \ugmcurr{} to predict recorded deaths among patients in hospice, inpatient, and ambulatory encounters. For hospice encounters, predictions are made at the start of a hospice admission encounter, with mortality assessed at 3, 7, 14, and 30 days. For inpatient encounters, predictions are made at the start of an inpatient-type encounter (inpatient admission, emergency department to inpatient, or hospital outpatient visit to inpatient), with mortality assessed at 7, 14, and 30 days. For ambulatory encounters, predictions are made at the end of an ambulatory-type encounter (office visits, walk-ins, immunizations, home care visits, allied health visits, occupational or physical therapy encounters, and routine prenatal encounters), with mortality assessed at 365 days. 

Index encounters are restricted to those occurring in 2022--2023 among patients aged 18 years or older. This eligibility window ensures at least 1 year of potential observation after the encounter, before the April 16, 2025 data cutoff. When multiple eligible encounters exist for a patient, one is randomly sampled. Model prompts contain only tokens available at prediction time. Hospice and inpatient prompts include prior history through the encounter-start token at the prediction instant to avoid same-time leakage, and ambulatory prompts include prior history through the encounter-end token.

For each prediction task, we sample 5,000 patients, oversampling positive cases to a minimum of 500 per outcome when available. We reweight samples to ensure prevalence-dependent metrics reflect the underlying population despite positive-case oversampling~\cite{curiosity_v1}. 

\begin{figure}[tbp]
    \centering
    \includegraphics[width=\linewidth]{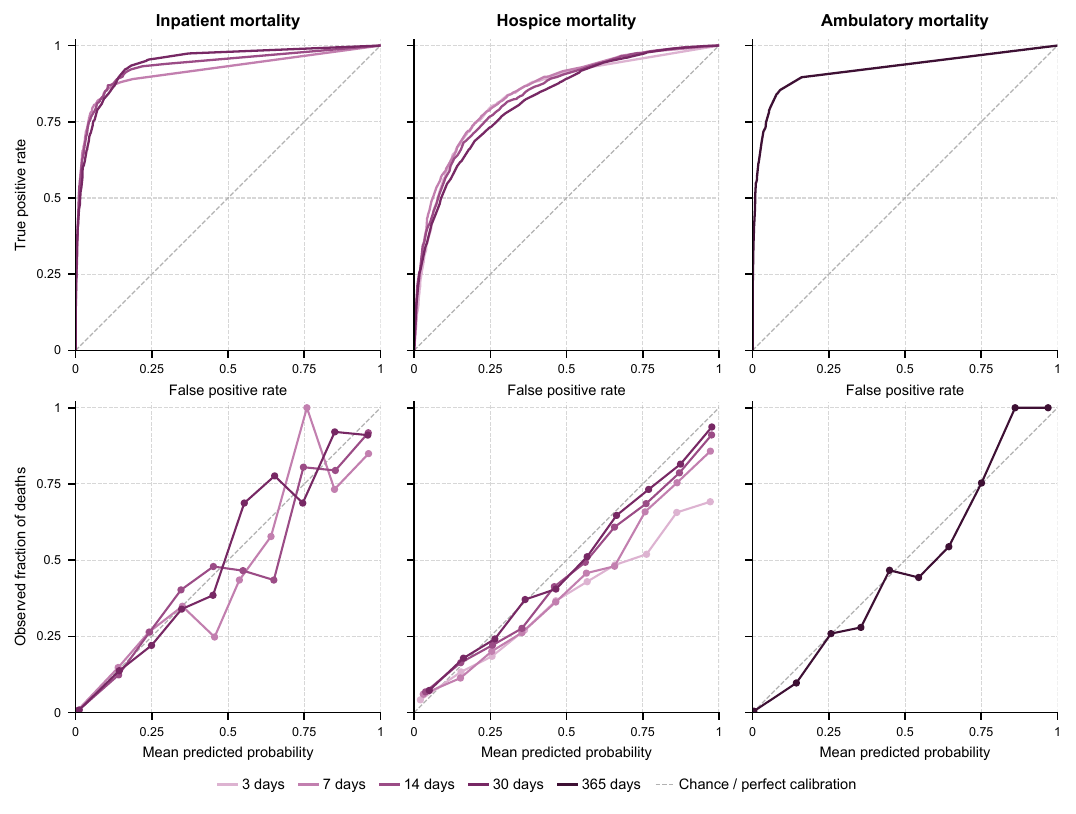}
    \caption[Mortality prediction performance]{\textbf{Mortality prediction performance and calibration.} Receiver operating characteristic and calibration curves for prediction of documented death. Panels show, left to right, inpatient (7, 14, and 30 days), hospice (3, 7, 14, and 30 days), and ambulatory (365 days) mortality windows. Calibration curves compare predicted mortality probabilities with observed mortality frequencies, reweighted to reflect the underlying population.}
    \label{fig:mortality_roc}
\end{figure}

We generate 60 simulations per patient of up to 3,000 tokens with four retries for ambulatory predictions; 60 simulations of up to 2,000 tokens with four retries for inpatient predictions; and 20 simulations of up to 2,000 tokens with three retries for hospice predictions. Simulations that neither contain the positive outcome nor reach the target follow-up time are considered incomplete. Incomplete simulations are extended through retry generation until they reach the requested generation horizon or terminate with an \ac{EOS} token. Simulations that remain incomplete after retries are discarded. Binary predictions at each mortality horizon are averaged to derive estimates of mortality probability, which are then compared with observed events. We assess calibration for each task by comparing predicted mortality probabilities with observed mortality frequencies. Calibration curves use the same sampling weights described above to account for positive-case oversampling. Results are shown in \autoref{fig:mortality_roc} and \autoref{tab:full_mortality}.

\subsubsection{Pediatric cohort evaluation tasks}
\label{subsubsec:eval-peds}

For pediatric patients, a population absent from \ugmprev{}, we stratify results into six age cohorts adapted from National Institute of Child Health and Human Development definitions~\cite{pedagegroups}: neonates ($<$28 days), infancy (28 days--12 months), toddler (13--23 months), early childhood (2--5 years), middle childhood (6--11 years), and early adolescence (12--17 years). We compare performance with a reference cohort of adults aged $\geq$18 years. Evaluation hyperparameters are held constant across age cohorts.

\textbf{Disease prediction.} We evaluate one incident disease prediction task and two acute-on-chronic disease prediction tasks among pediatric patients over a two-year prediction window. We exclude neonate, infant, and toddler cohorts from the disease prediction tasks, as these outcomes are too rare in these age cohorts to estimate performance reliably. Full phenotype definitions are described in \citet{curiosity_v1}. 
\begin{itemize}
  \item \textbf{Incident asthma.} Among patients with no prior qualifying asthma diagnosis and at least two prior outpatient or acute encounters (inpatient admission, emergency department, or urgent care) in the preceding two years, we predict whether they will meet asthma phenotype criteria within the next two years. The outcome is defined by a qualifying \ac{ICD-10-CM} code documented in at least two outpatient encounters or in a single acute encounter, using encounter and billing diagnoses only. Prediction occurs at the end of a randomly selected outpatient encounter.   
  \item \textbf{Asthma exacerbation.} Among patients with established asthma, we predict occurrence of asthma exacerbation during an acute encounter. Asthma must be documented before prediction in at least two outpatient encounters or one acute encounter; patients with prior acute events remain eligible. 
  \item \textbf{Sickle cell crisis.} Among patients with established sickle cell disease, we predict occurrence of sickle cell crisis during an acute encounter. Sickle cell disease must be documented before prediction in at least two outpatient encounters or one acute encounter; patients with prior acute events remain eligible. 
\end{itemize}  

\begin{figure}[tbp]
    \centering
    \includegraphics[width=\linewidth]{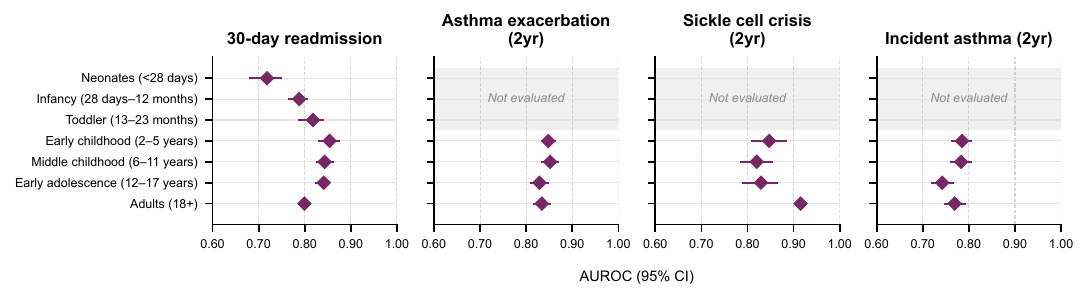}
    \caption[Pediatric cohort outcome predictions]{\textbf{Predictions for 30-day readmission and selected disease events, by pediatric age group.} AUROC with 95\% confidence intervals for 30-day readmission, asthma exacerbation, sickle cell crisis, and incident asthma, shown for each pediatric age cohort and the adult reference cohort. Disease events are predicted over a two-year window; patients younger than 2 years were not evaluated for these tasks.}
    \label{fig:age_cohort_forest}
\end{figure}

For each task and age cohort, we sample 5{,}000 patients, oversampling positive cases to a minimum of 500 per outcome when available and applying inverse-resampling weights to metrics. We generate 60 simulations up to 4{,}000 tokens per patient, with up to two retries for any simulation that does not reach the two-year horizon. Simulations that remain incomplete are discarded. Results are shown in \autoref{fig:age_cohort_forest}.

\textbf{30-day readmission after pediatric encounters.} At inpatient discharge, we predict occurrence of a new inpatient encounter within 30 days. Readmissions occurring within 24 hours of discharge are excluded, as they often reflect interfacility transfers or erroneous discharge events. We sample up to 10{,}000 encounters per age cohort, with more than 3{,}000 in the smallest cohort. We generate 80 simulations of up to 1{,}500 tokens for each sampled encounter, discarding any simulation that neither contains a readmission event nor spans the 30-day window. Results are shown in \autoref{fig:age_cohort_forest}.

\begin{figure}[tbp]
    \centering
    \includegraphics[width=\linewidth]{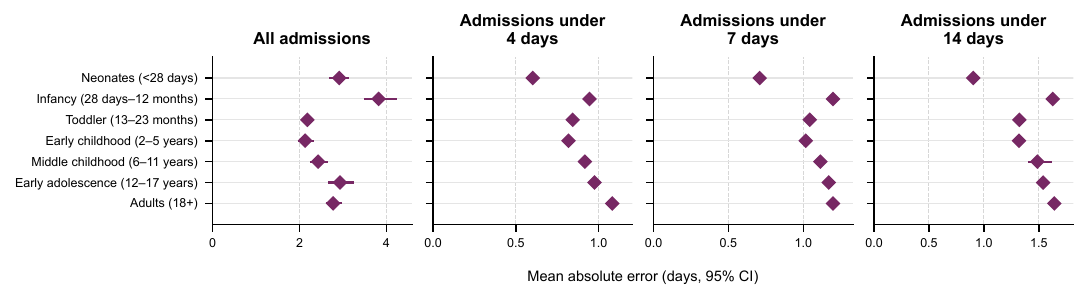}
    \caption[Pediatric length of stay predictions]{\textbf{Length-of-stay prediction error for pediatric admissions, by age group and admission duration.} Mean absolute error in days with 95\% confidence intervals for each pediatric age cohort and the adult reference cohort, shown for all admissions and admissions shorter than 4, 7, and 14 days.}
    \label{fig:age_cohort_los_forest}
\end{figure}

\textbf{Length of stay for pediatric admissions.} At inpatient admission, we predict time to discharge. The model is prompted with the patient's history through the admission encounter header, which includes encounter type, department specialty, and any chief complaints. The flagged primary encounter diagnosis, which was included in the prompt for \ugmprev{}, is not provided to \ugmcurr{}. We sample 10{,}000 encounters per age cohort, generating 20 inpatient simulations of up to 3{,}000 tokens each per encounter. Simulations that do not include an encounter-end token are discarded. The predicted length of stay is the median simulated time to encounter end. \autoref{fig:age_cohort_los_forest} shows \ac{MAE} in days for all admissions and for admissions shorter than 4, 7, and 14 days. 

\begin{figure}[tbp]
    \centering
    \includegraphics[width=\textwidth]{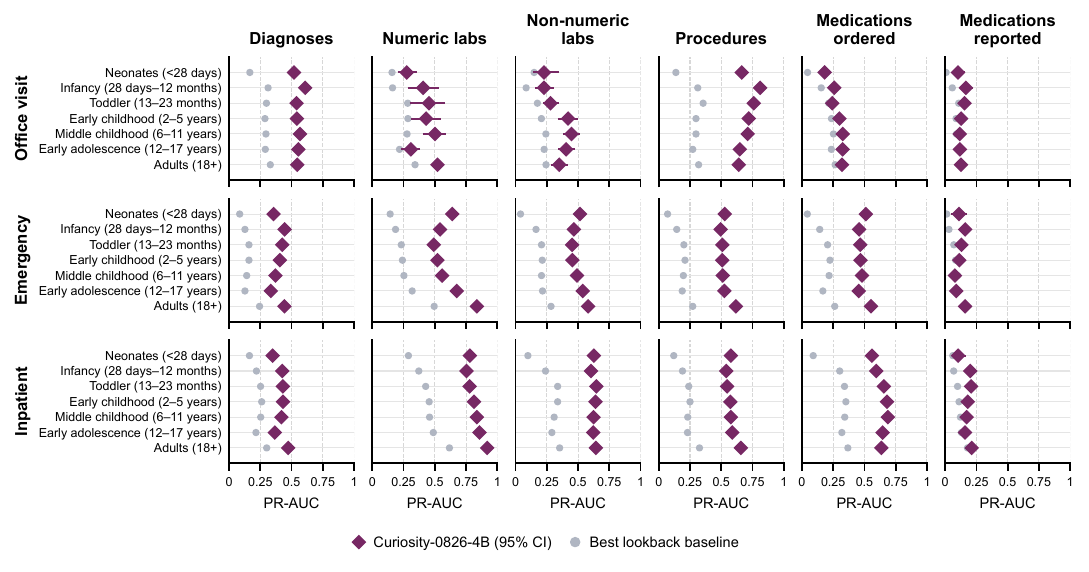}
    \caption[Predicted events within single encounters among pediatric patients]{\textbf{Predicted events within single encounters among pediatric patients.} \ac{PR-AUC} across encounter types (office visit, emergency, and inpatient) and event types (diagnoses, numeric and non-numeric laboratory tests collected, procedures, and ordered and reported medications) with 95\% confidence intervals for each pediatric age cohort alongside the adult reference cohort. Gray points indicate the best lookback baseline for that cohort and task, with lookback windows ranging from one month to the patient's full history.}
    \label{fig:aucprc_dotgrid}
\end{figure}

\textbf{Single-encounter generation for pediatric settings.} We predict unique medical events occurring within individual office visits, emergency department visits, and inpatient admissions. Predicted event types include diagnoses, numeric and non-numeric laboratory tests collected, ordered and reported medications, and procedures. For each encounter type and age cohort, we sample 5{,}000 random encounters without requiring a minimum amount of prior history. Prompts end after the encounter header, and we generate 20 simulations of up to 4{,}000 tokens per encounter until the model emits the matching encounter end token. Simulations that do not contain an encounter end token are discarded. A correctly predicted event within an encounter must exactly match one documented in the real encounter, including for multi-token diagnosis and medication events. We compute micro-averaged precision and recall across thresholds and report \ac{PR-AUC}. We compare model predictions with a lookback baseline, which uses previously documented events to predict the single-encounter events. We evaluate eight lookback windows: 1, 2, 3, and 6 months; 1, 2, and 3 years; and the patient's full history. For each age cohort, encounter type, and event type, we report the highest PR-AUC achieved across these windows. Results are shown in \autoref{fig:aucprc_dotgrid}.

\subsection{Comparison with \ugmprev{} on existing capabilities}
\label{subsec:eval-existing}

We compare \ugmcurr{} with \ugmprev{} in adult patients (aged $\geq$18 years) on selected tasks: 30-day readmission, length of stay, laboratory result prediction, incident disease prediction, and single-encounter generation. The largest numerical improvements with \ugmcurr{} are in laboratory result prediction, with smaller changes across the other tasks. Shared evaluation tasks were a representative subset of previously evaluated tasks~\cite{curiosity_v1} that span short and long time horizons, numeric and categorical outcomes, and clinical and operational tasks. 

Cohort eligibility, sampling procedures, prediction points, outcome definitions, simulation aggregation, and scoring follow methods in~\citet{curiosity_v1}. Inference settings, including simulation counts, generation token limits, retries, and handling of incomplete simulations, also follow previously described methods. \ugmcurr{} receives a larger token budget than \ugmprev{} throughout, because the added data types increase the average number of tokens per patient, and more tokens are required to cover equivalent calendar time. Our previous study showed that predictive performance improves as more simulations are used to produce each prediction~\cite{curiosity_v1}. Therefore, we use the same number of simulations per prediction across cohorts within each task. In each comparison between models, \ugmprev{} uses at least as many simulations per prediction as \ugmcurr{}, so any improvement with \ugmcurr{} cannot be attributed to higher simulation counts. 

\subsubsection{30-day readmission after adult encounters}
\label{subsubsec:eval-readmission}

\begin{figure}[tbp]
    \centering
    \includegraphics[width=110mm]{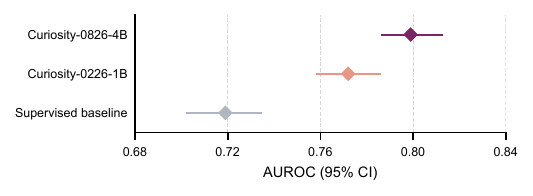}
    \caption[30-day readmission \ac{AUROC}]{\textbf{Prediction of 30-day readmission for adult patients.} \ac{AUROC} with 95\% confidence intervals for predicting hospital readmission within 30 days of discharge across 10{,}000 hospital encounters. \ugmcurr{} achieved an \ac{AUROC} of 0.799 compared with 0.772 for \ugmprev{} and 0.719 for a supervised baseline. The axis begins at 0.68.}
    \label{fig:readmission_roc}
\end{figure}

We predict hospital readmission within 30 days of discharge following methods described in \autoref{subsubsec:eval-peds}. Evaluation includes 10{,}000 encounters per model. \ugmcurr{} achieved an \ac{AUROC} of 0.799 compared with 0.772 for \ugmprev{}. A supervised baseline, described in \citet{curiosity_v1}, is included for comparison. Results are shown in \autoref{fig:readmission_roc}.

\subsubsection{Length of stay for adult admissions}
\label{subsubsec:eval-los}

The length of stay evaluation for both models follows methods described in \autoref{subsubsec:eval-peds}, except that \ugmprev{} generates up to 2{,}000 tokens rather than 3{,}000. 

\begin{figure}[tbp]
    \centering
    \includegraphics[width=0.5\textwidth]{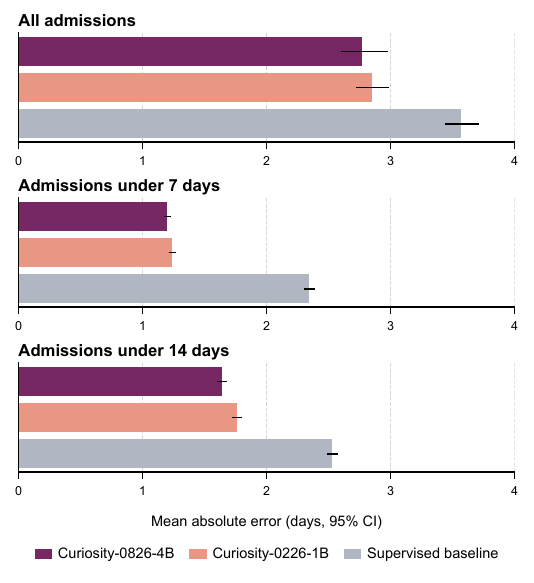}
    \caption[Length-of-stay prediction error]{\textbf{Length-of-stay prediction error for adult admissions, by admission duration.} Mean absolute error (MAE) in days with 95\% confidence intervals, shown for all admissions and admissions lasting <7 and <14 days, for \ugmcurr{} and \ugmprev{}. A supervised baseline is shown for comparison. MAE decreased by 0.08 days for all admissions, 0.04 for admissions <7 days, and 0.12 for admissions <14 days using \ugmcurr{} compared with \ugmprev{}. 
    }
    \label{fig:los_mae}
\end{figure}

Using \ugmcurr{}, \ac{MAE} decreased from 2.85 to 2.77 days across all admissions. For admissions shorter than 14 days, it decreased from 1.76 to 1.64 days, and for admissions shorter than 7 days, it decreased from 1.24 to 1.20 days. A supervised baseline is included for comparison~\cite{curiosity_v1}. Results are shown in \autoref{fig:los_mae} and \autoref{tab:full_los}.

\subsubsection{Laboratory value predictions}
\label{subsubsec:eval-labs}

We predict \ac{HgbA1c} value for patients treated for \ac{T2DM}, and \ac{HDL}, \ac{LDL}, and total cholesterol for patients treated for hyperlipidemia 60 to 120 days after a change in treatment for the condition. Cohort construction, medication order prediction points, and sampling of continuous predicted values from generated laboratory value buckets follow methods described in \citet{curiosity_v1}. Briefly, the prompt begins at a patient's first token and continues through a randomly selected new or changed order for medication treating the underlying condition. 

\ugmcurr{} generates 64 simulations per sample of up to 4{,}000 tokens each, while \ugmprev{} generates 80 simulations of up to 2{,}000 tokens. We discard simulations that do not cover the window from 60 to 120 days following the index date. From the retained simulations, we parse the laboratory values generated within the target window and compare them against the ground truth. When a simulation contains more than one laboratory value of the target type within the window, we average those samples. Aggregate performance for each laboratory test type uses between 3{,}000 and 6{,}000 samples per model. 

\begin{figure}[tbp]
    \centering
    \includegraphics[width=0.5\textwidth]{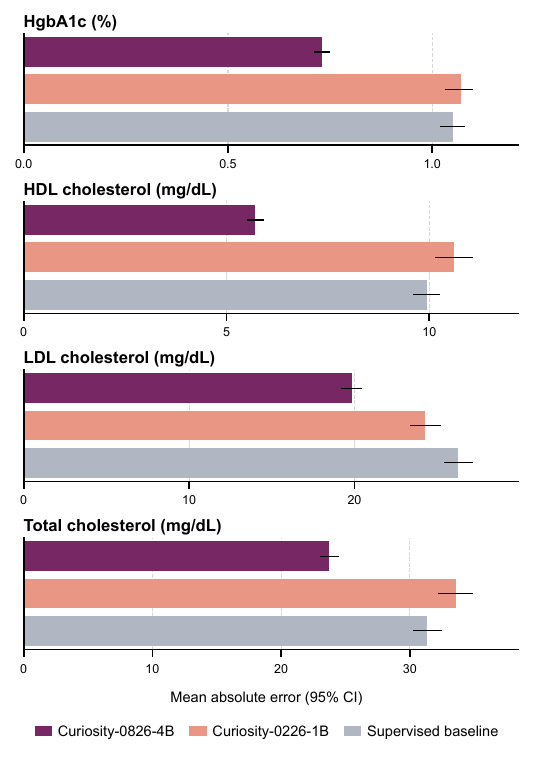}
    \caption[Lab prediction error]{\textbf{Laboratory value prediction error in adults.} Mean absolute error (MAE) with 95\% confidence intervals for laboratory value predictions 60 to 120 days after a change in medication treatment for \ac{T2DM} (HgbA1c) or hyperlipidemia (HDL, LDL, total cholesterol), comparing \ugmcurr{} with \ugmprev{} and a supervised baseline. MAE decreased by 31.8\% for HgbA1c, 46.2\% for HDL, 18.3\% for LDL, and 29.3\% for total cholesterol using \ugmcurr{} compared with \ugmprev{}.}
    \label{fig:lab_mae}
\end{figure}

\ac{MAE} decreased with \ugmcurr{} compared with \ugmprev{} across the laboratory tests evaluated, with relative reductions of 18 to 46\%. A supervised baseline, as described in \citet{curiosity_v1}, is included for comparison. Results are shown in \autoref{fig:lab_mae} and \autoref{tab:full_labs}. 

\subsubsection{Incident disease prediction in adults}
\label{subsubsec:eval-incident}

We evaluate two-year incident disease prediction for \ac{COPD}, \ac{CHF}, dementia, asthma, alcohol use disorder, and acute myocardial infarction. Cohort eligibility and phenotype definitions are described in \citet{curiosity_v1}. Sampling, weighting, and inference follow methods described for the pediatric evaluation in \autoref{subsubsec:eval-peds}. For \ugmcurr{}, up to 4{,}000 tokens are generated per simulation, compared with up to 2{,}000 for \ugmprev{}, to cover equivalent calendar time. Multiple incident disease targets are evaluated from each simulation. 

\begin{figure}[tbp]
    \centering
    \includegraphics[]{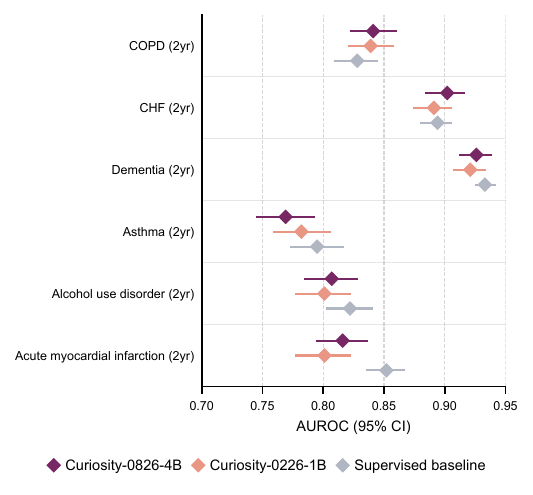}
    \caption[Incident disease \ac{AUROC}]{\textbf{Incident disease prediction in adults.} \ac{AUROC} with 95\% confidence intervals for two-year prediction of incident \ac{COPD}, \ac{CHF}, dementia, asthma, alcohol use disorder, and acute myocardial infarction, comparing \ugmcurr{}, \ugmprev{}, and a supervised baseline. The axis begins at 0.70.}
    \label{fig:incident_auroc_diff}
\end{figure}

Using \ugmcurr{}, \ac{AUROC} across tasks was comparable to that of \ugmprev{}, with slight increases seen for  \ac{COPD}, \ac{CHF}, dementia, alcohol use disorder, and acute myocardial infarction (+0.2--1.9\%) and a slight decrease for asthma (-1.7\%) (\autoref{fig:incident_auroc_diff}). A supervised baseline is shown for comparison \cite{curiosity_v1}. 

\subsubsection{Single-encounter generation for adults}
\label{subsubsec:eval-single-encounter}

We predict unique events occurring within single office visits, emergency department encounters, and inpatient encounters. \ugmprev{} samples 10{,}000 encounters of each encounter type and generates up to 2{,}000 tokens per simulation; \ugmcurr{} samples 5{,}000 encounters of each type and generates up to 4{,}000 tokens per simulation. Encounter eligibility, incomplete sequence handling, and target scoring rules otherwise follow methods described in \autoref{subsubsec:eval-peds}. Predictions for non-numeric laboratory tests and reported medications are evaluated for \ugmcurr{} only, as these capabilities were not present for \ugmprev{}. 

\begin{figure}[tbp]
    \centering
    \includegraphics[width=\linewidth]{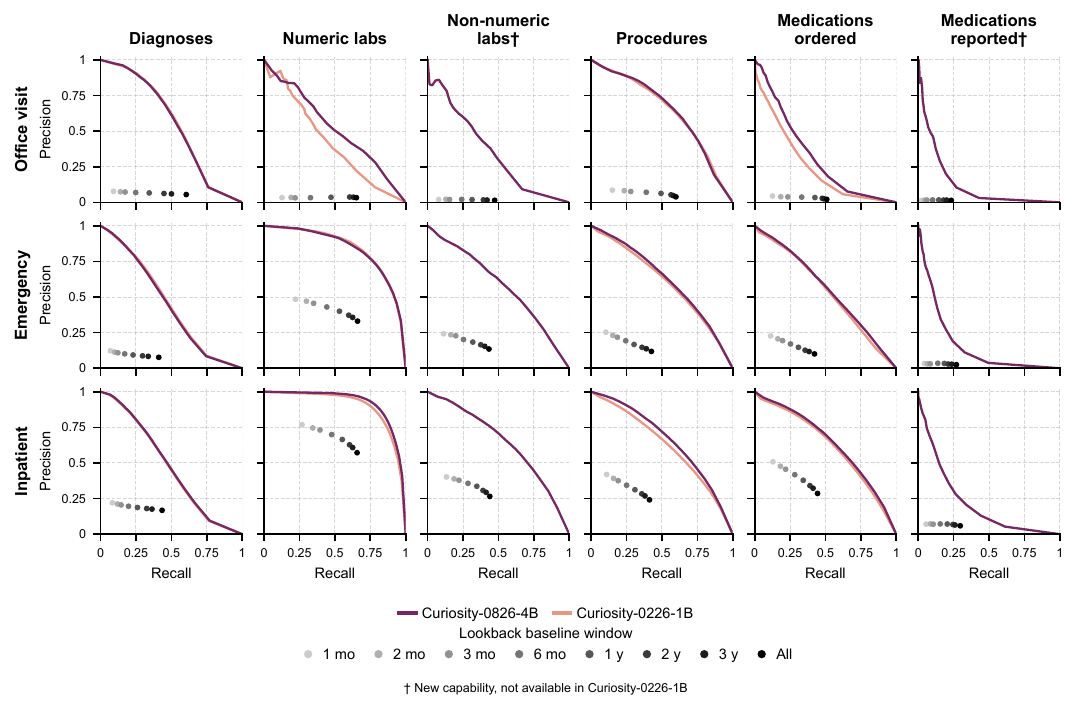}
    \caption[single-encounter generation performance]{\textbf{Predicted events within single encounters among adult patients.} Precision-recall curves for \ugmcurr{} and \ugmprev{} by encounter type (office visit, emergency, and inpatient) and event type (diagnoses, numeric and non-numeric laboratory tests collected, procedures, and ordered and reported medications), among adult patients. Non-numeric laboratory tests and reported medications (†) are new in \ugmcurr{} and have no \ugmprev{} curve. Gray points show the lookback baselines for each task using lookback periods of 1, 2, 3, and 6 months; 1, 2, and 3 years; and the patient's entire history.}
    \label{fig:el_prauc}
\end{figure}

\autoref{fig:el_prauc} shows precision-recall curves by encounter type and event targets; \ugmprev{} is shown for comparison where applicable. \autoref{tab:full_encounter_gen} shows PR-AUCs and 95\% CIs. Overall, performance was broadly comparable across tasks shared with \ugmprev{}, with improvements in several event categories. The largest improvements with \ugmcurr{} were seen in predicting collected numeric laboratory tests (0.421 to 0.521) and ordered medications (0.251 to 0.320) for office visits. In inpatient encounters, PR-AUC increased across all events using \ugmcurr{}, with the largest improvement in prediction of procedures (0.619 to 0.657) and collected numeric laboratory tests (0.899 to 0.919). 

\ugmcurr{} supports prediction of two new targets: non-numeric laboratory tests and reported medications. Non-numeric laboratory tests demonstrated the highest PR-AUCs in inpatient and emergency department encounters (0.643 and 0.581, respectively) compared with 0.350 for office visits. PR-AUCs for reported medications were relatively low across encounter types (0.127--0.212). 

\section{Discussion}
\label{sec:discussion}

\ugmcurr{} expands the scope of zero-shot clinical prediction while generally matching or exceeding \ugmprev{} on evaluated tasks supported by both models \cite{curiosity_v1}. Broader population coverage and richer event representations bring pediatric prediction tasks and additional outcome types, including documented mortality, into the common generative framework. This expansion provides a broader foundation for developing clinical and operational applications that could benefit patients and health systems without requiring separate task-specific models.

\subsection{Richer representations and new capabilities}
\label{subsec:disc-richer}
Clinical and operational planning often requires considering several possible outcomes for the same patient. Other models have explored generative forecasting of patient timelines~\cite{foresight, ETHOS, delphi}. Curiosity-0826-4B extends our earlier framework~\cite{curiosity_v1}, creating more opportunities to investigate multiple related outcomes with a single model. Examining multiple outcomes within a common simulated timeline aligns better with real decision making; for example, clinical reasoning considers disease progression, markers of severity, and likely care needs together. This capability could provide a more comprehensive basis for anticipating healthcare needs. 

\ugmcurr{} extends the cohort of patients used in training to include pediatric patients. By incorporating pediatric patients and events related to birth, as well as adapting the representation of existing data types to better reflect pediatric data, \ugmcurr{} supports prediction tasks across younger age cohorts that benefit from pediatric-specific representations. Performance on most shared evaluation tasks was comparable between eligible pediatric cohorts and the adult reference population, although sickle cell crisis AUROCs were lower in pediatric cohorts than in adults, and 30-day readmission AUROC was lowest among neonates. For pediatric patients, understanding future disease-related events may support more personalized communication with patients and families about disease risks, earlier recognition of disease, more targeted interventions and preventive strategies, and more informed planning for healthcare utilization. Pediatric care encompasses changing physical, developmental, and psychosocial needs and the need for thoughtful transition to adult care \cite{hardin2017}. A single model trained across pediatric and adult populations provides a common framework for forecasting across life stages, without requiring separate age-specific models. This creates an opportunity to investigate greater continuity in prediction as patients transition from pediatric to adult care.

Beyond extending population coverage, richer event representations broaden both the clinical information available to the model and the outcomes it can predict. While the larger model showed improvement in many tasks shared with the previous model, it did not yield uniform improvements across shared evaluation tasks. The greater gains in laboratory result prediction are consistent with task-specific benefits from finer value resolution, revised laboratory result timing, and additional clinical context, and prior work has shown benefits of finer numerical representations \cite{labtop}. However, current comparisons do not isolate these contributions from changes in model scale and training. Representation changes also expand the range of questions the model can address: adding documented death enables mortality prediction, while explicit right-censoring representations distinguish the end of observed follow-up from a recorded death. Mortality risk scores have been used to identify patients for proactive care planning, including enrollment in home-based care, discussing prognosis and goals of care, and documenting end-of-life preferences \cite{joseph2026}. These examples illustrate how improved event representation can support more detailed prediction within existing tasks and expansion to new outcome types.

Across a range of clinical and operational applications, understanding how to combine predictions about event occurrence and timing for different users, settings, and needs requires deep knowledge of workflow. Once validated for these purposes, a single model with broad capabilities could inform clinicians’ diagnostic decisions and support resource allocation and capacity planning. These potential benefits depend on understanding which outcomes should be considered together and how to present them to the people making those decisions. 

\subsection{Limitations and future areas of work}
\label{subsec:disc-limitations}
As with the previous work \cite{curiosity_v1}, the evaluations in this report are retrospectively constructed from historical EHR data. The outcomes evaluated here are defined by phenotypes in documentation, which can vary in how they correspond to the clinical events they are intended to capture. This limitation remains particularly important for mortality estimation, as deaths not documented in the EHR are under-ascertained. While the current model explicitly adds representation for right-censoring, the underlying timeline remains an imperfect representation of a patient's health trajectory. Evaluation tasks, such as predicting length of stay, excluded simulations that failed to complete within the maximum sequence length. This disproportionately affects sequences covering longer calendar time and those with high density of documented events, limiting the representativeness of retained simulations.

While significant progress has been made, several gaps remain before routine clinical use is feasible. The retrospective performance reported here does not by itself establish that simulation-derived estimates are calibrated individual clinical risks. Prospective evaluation studies are needed to assess model performance in unseen workflows, settings, and temporal ranges, as well as validation and calibration in local settings and clinically meaningful subgroups \cite{Bedi2026ZeroShot}. While pediatric age subgroups were assessed in this study, other aggregations may obscure differences across clinically important patient groups. Clinical deployment also requires understanding how predictions will be integrated into clinical and operational workflows in ways that are timely, actionable, and beneficial to patient care \cite{FURM}. As with other models derived from observational data, predicting observed events after documented treatment changes is not equivalent to estimating the causal effect of changing care, underscoring the need to evaluate downstream outcomes. On the technical front, as patient timeline representations grow more detailed, efficient methods for handling larger context, tokenization, and inference remain important areas for future development, particularly for simulation over longer time horizons. 

\ugmcurr{} creates a broader foundation for exploring clinical and operational questions using generative medical event simulation, incorporating broader populations and richer timeline representations while generally matching or exceeding \ugmprev{} on evaluated tasks. Realizing this potential will require validation and calibration in intended populations and settings. Its ultimate impact will depend on thoughtfully connecting model simulations to actionable decisions, and on showing through prospective validation that resulting applications improve care and planning for patients and health systems. 
\bibliographystyle{unsrtnat}
\bibliography{references}

\clearpage
\appendix
\titleformat{\section}{\normalfont\Large\bfseries}{Appendix \thesection:}{.5em}{}

\section{\modelname{} tokenization}
\label{sec:token_changes}

\begin{figure}[H]
  \centering
\includegraphics[
  width=140mm,
  keepaspectratio
]{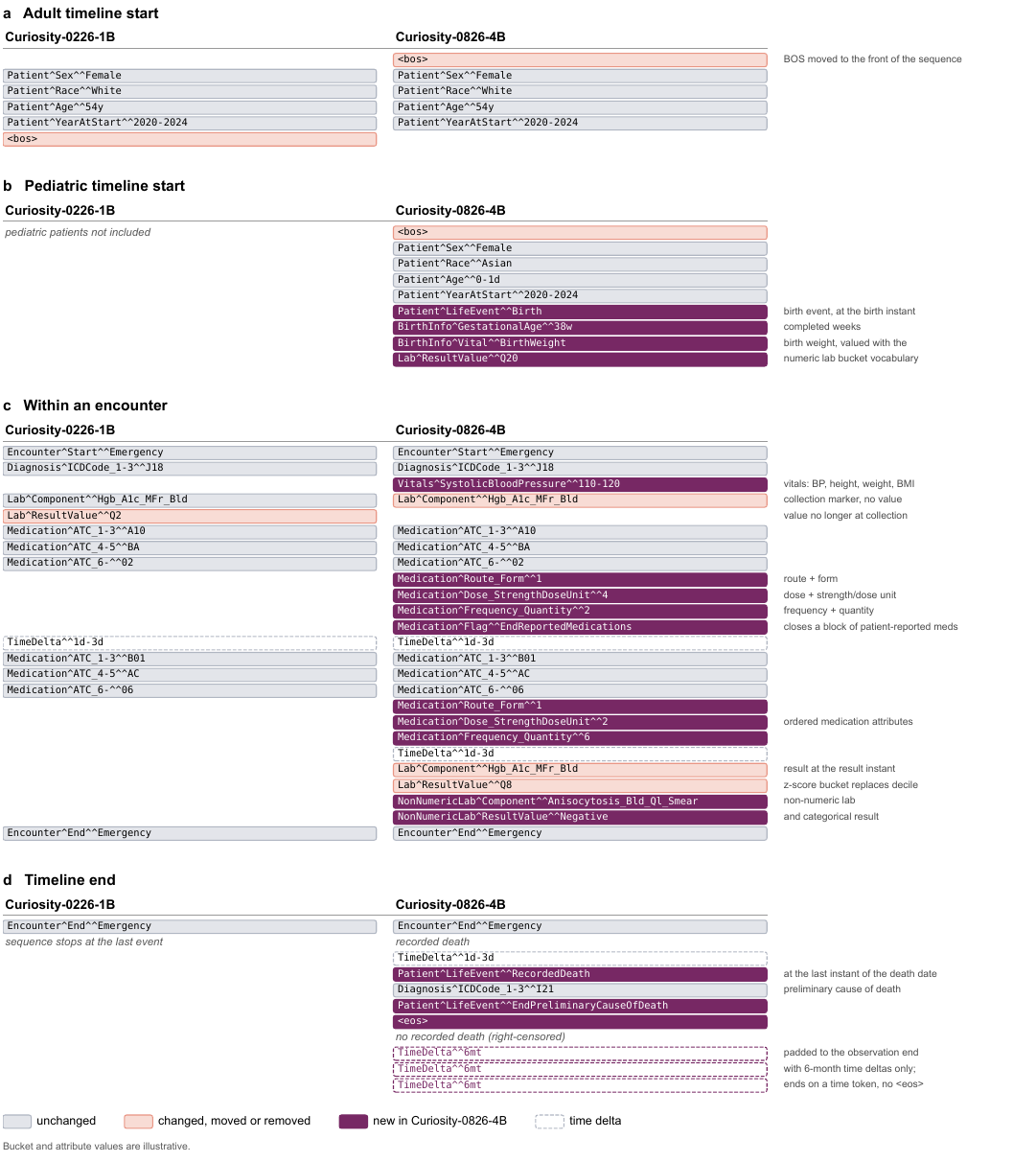}
  \caption[Tokenization changes from \ugmprev{} to \ugmcurr]{\textbf{Tokenization changes from \ugmprev{} to \ugmcurr{}.}
    Example token sequences under \ugmprev{} (left) and \ugmcurr{} (right). Bucket and attribute values are illustrative.
    (a)~\acs{BOS} moves to the front of the sequence.
    (b)~Pediatric patients are included, with birth, gestational age, and birth weight tokens at the birth instant.
    (c)~Vitals are added. Laboratory tests emit a value-less collection marker at collection and the result at the result instant, in z-score buckets instead of deciles. Non-numeric laboratory tests are added. Medication events gain route/form, dose/strength and frequency/quantity tokens, and patient-reported medications are closed by a flag.
    (d)~Recorded deaths emit a death token, the preliminary cause of death, and \acs{EOS}. For patients with no recorded death, timelines are padded after the last recorded medical event with 6-month time deltas. Expanded procedure coverage does not change any token formats and is not shown. See \autoref{sec:tokenization}.}
  \label{fig:tokenization_changes}
\end{figure}

\begin{table}[H]
  \centering
  \small
  \setlength{\tabcolsep}{4pt}
  \renewcommand{\arraystretch}{1.15}
  \begin{tabularx}{\textwidth}{@{} l >{\raggedright\arraybackslash}X >{\raggedright\arraybackslash}X >{\raggedright\arraybackslash}X c @{}}
    \toprule
    \textbf{Vital (unit)} & \textbf{Low end} & \textbf{Middle range} & \textbf{High end} & \textbf{Bins} \\
    \midrule
    Systolic BP (mmHg)            & 0--80          & 80--140 (by 10) & 140--200 (by 20), 200--370 & 13 \\
    Diastolic BP (mmHg)           & 0--50          & 50--120 (by 10) & 120--360                   & 11 \\
    Height (in)                   & 0--10          & 10--100 (by 5)  & 100--108                   & 22 \\
    Weight (lb)                   & 0--50 (by 1)   & 50--100 (by 2), 100--160 (by 3), 160--260 (by 5) & 260--300 (by 10), 300--480 (by 20), above 480 & 130 \\
    BMI (kg/m\textsuperscript{2}) & 0--18          & 18--40 (by 1)   & 40--204                    & 26 \\
    \bottomrule
  \end{tabularx}
  \caption[Vitals tokenization bins]{\textbf{Vital sign bins.} Each vital is emitted as a single token that includes the bin for the value: \tok{Vitals\cx<type>\cx\cx<bin>} (\autoref{subsec:tok-vitals}). Bins are narrow across the clinically common range, with extra resolution in ranges relevant to pediatric patients, and wider toward the extremes; ``by 10'' means one bin per 10 units. Bins are left-edge exclusive, so a value on a boundary falls in the lower bin (a systolic pressure of 120~mmHg falls in 110--120). Values outside the plausibility bounds are removed before binning. The bin count includes one catch-all bin below the lowest threshold and one above the highest.}
  \label{tab:vitals_bins}
\end{table}

\clearpage
\section{\modelname{} dataset statistics}
\label{sec:dataset_stats}

After the pre-processing described in \autoref{sec:tokenization}, data from Epic Cosmos is transformed into the subset used to train \modelname{}, split into train and test sets at the patient level. Summary statistics of this dataset are shown in \autoref{tab:supp_datset_statistics}.

\begin{table}[H]
  \centering
  \begin{tabular}{llrrr}
    \toprule
    \textbf{Characteristic} & \textbf{Group} &
    \textbf{Train} & \textbf{Test} &
    \textbf{Total} \\
    \midrule
    \multirow{13}{*}{Total counts}
      & Patients          & 172 M   & 19.1 M & 191 M  \\
      & Encounters        & 9.87 B  & 1.10 B & 11.0 B \\
      & Diagnoses         & 18.8 B  & 2.09 B & 20.9 B \\
      & Numeric labs      & 28.4 B  & 3.15 B & 31.5 B \\
      & Non-numeric labs  & 5.36 B  & 596 M  & 5.96 B \\
      & Medications       & 8.49 B  & 943 M  & 9.43 B \\
      & Procedures        & 14.9 B  & 1.65 B & 16.5 B \\
      & Vitals            & 12.5 B  & 1.39 B & 13.9 B \\
      & Births            & 12.6 M  & 1.40 M & 14.0 M \\
      & Deaths            & 5.46 M  & 606 K  & 6.07 M \\
      & Tokens            & 289 B   & 32.1 B & 321 B  \\
      & Tokens/patient    & 1685.0  & 1684.4 & 1684.9 \\
      & Tokens/encounter  & 29.3    & 29.3   & 29.3   \\
    \midrule
    \multirow{10}{*}{Age}
      & Neonates ($<$28 days)            & 11.3 M & 1.25 M & 12.5 M \\
      & Infancy (28 days--12 months)     & 3.47 M & 384 K  & 3.86 M \\
      & Toddler (13--23 months)      & 2.20 M & 246 K  & 2.45 M \\
      & Early childhood (2--5 years)     & 6.67 M & 740 K  & 7.41 M \\
      & Middle childhood (6--11 years)   & 9.39 M & 1.04 M & 10.4 M \\
      & Early adolescence (12--17 years) & 11.0 M & 1.22 M & 12.2 M \\
      & 18--39 years                     & 49.1 M & 5.45 M & 54.6 M \\
      & 40--59 years                     & 41.8 M & 4.65 M & 46.5 M \\
      & 60--79 years                     & 30.5 M & 3.39 M & 33.9 M \\
      & 80+ years                        & 6.06 M & 673 K  & 6.73 M \\
    \midrule
    \multirow{7}{*}{Race}
      & White       & 105 M  & 11.7 M & 117 M  \\
      & BoAA        & 24.3 M & 2.70 M & 27.0 M \\
      & Asian       & 7.27 M & 807 K  & 8.08 M \\
      & AI/AN       & 1.76 M & 197 K  & 1.96 M \\
      & NHOPI       & 720 K  & 79.6 K & 799 K  \\
      & Other       & 8.69 M & 966 K  & 9.65 M \\
      & Unspecified & 23.3 M & 2.58 M & 25.8 M \\
    \midrule
    \multirow{6}{*}{Sex}
      & Female      & 91.9 M & 10.2 M & 102 M  \\
      & Male        & 79.6 M & 8.84 M & 88.4 M \\
      & Unknown     & 56.4 K & 6.33 K & 62.7 K \\
      & Masked      & 11.3 K & 1.26 K & 12.5 K \\
      & Other       & 2.25 K & 248    & 2.50 K \\
      & Unspecified & 1.07 K & 90     & 1.16 K \\
    \bottomrule
  \end{tabular}
  \caption[\modelname{} dataset statistics]{\modelname{} dataset counts in the train, test, and full data sets, organized by various data types and demographics. Abbreviations: BoAA = Black or African American, AI/AN = American Indian or Alaska Native, NHOPI = Native Hawaiian or Other Pacific Islander. Race reflects the first listed value.}
  \label{tab:supp_datset_statistics}
\end{table}

\section{Tabulated results}
\label{sec:full_results}

The tables below report the full numeric results behind the figures in \autoref{sec:evaluation}, with 95\% confidence intervals. 

\begin{table}[H]
  \centering
  \small
  \setlength{\tabcolsep}{6pt}
  \begin{tabular}{llc}
    \toprule
    & & \textbf{AUROC} \\
    \textbf{Cohort} & \textbf{Horizon} & \textbf{(95\% CI)} \\
    \midrule
    \multirow{4}{*}{Hospice}
      & 3 days   & \makecell[c]{0.837 \\ {\scriptsize{(0.822, 0.852)}}} \\
      & 7 days   & \makecell[c]{0.848 \\ {\scriptsize{(0.836, 0.859)}}} \\
      & 14 days  & \makecell[c]{0.837 \\ {\scriptsize{(0.826, 0.848)}}} \\
      & 30 days  & \makecell[c]{0.821 \\ {\scriptsize{(0.809, 0.832)}}} \\
    \midrule
    \multirow{3}{*}{Inpatient}
      & 7 days   & \makecell[c]{0.916 \\ {\scriptsize{(0.899, 0.932)}}} \\
      & 14 days  & \makecell[c]{0.932 \\ {\scriptsize{(0.918, 0.946)}}} \\
      & 30 days  & \makecell[c]{0.942 \\ {\scriptsize{(0.931, 0.952)}}} \\
    \midrule
    \multirow{1}{*}{Ambulatory}
      & 365 days & \makecell[c]{0.921 \\ {\scriptsize{(0.905, 0.936)}}} \\
    \bottomrule
  \end{tabular}
  \caption[Mortality prediction results]{\textbf{Mortality prediction.} \ac{AUROC} with 95\% confidence intervals for prediction of documented death within relevant mortality windows for hospice, inpatient, and ambulatory cohorts (\autoref{subsubsec:eval-mortality}). These are the tabulated values behind \autoref{fig:mortality_roc}.}
  \label{tab:full_mortality}
\end{table}

\begin{table}[H]
  \centering
  \small
  \setlength{\tabcolsep}{6pt}
  \begin{tabular}{lcccc}
    \toprule
    &  \textbf{30-day} & \textbf{Asthma} & \textbf{Sickle cell} & \textbf{Incident} \\
        \textbf{Age cohort} & \textbf{readmission} & \textbf{exacerbation} & \textbf{crisis} & \textbf{asthma} \\
    \midrule
    Neonates ($<$28 days) & \makecell[c]{0.718 \\ {\scriptsize{(0.679, 0.750)}}} & & & \\
    Infancy (28 days--12 months) & \makecell[c]{0.788 \\ {\scriptsize{(0.763, 0.807)}}} & & & \\
    Toddler (13--23 months) & \makecell[c]{0.818 \\ {\scriptsize{(0.785, 0.842)}}} & & & \\
    Early childhood (2--5 years) & \makecell[c]{0.854 \\ {\scriptsize{(0.829, 0.876)}}} & \makecell[c]{0.848 \\ {\scriptsize{(0.832, 0.865)}}} & \makecell[c]{0.847 \\ {\scriptsize{(0.807, 0.885)}}} & \makecell[c]{0.785 \\ {\scriptsize{(0.762, 0.807)}}} \\
    Middle childhood (6--11 years) & \makecell[c]{0.843 \\ {\scriptsize{(0.824, 0.863)}}} & \makecell[c]{0.852 \\ {\scriptsize{(0.833, 0.871)}}} & \makecell[c]{0.820 \\ {\scriptsize{(0.783, 0.855)}}} & \makecell[c]{0.783 \\ {\scriptsize{(0.758, 0.807)}}} \\
    Early adolescence (12--17 years) & \makecell[c]{0.841 \\ {\scriptsize{(0.823, 0.855)}}} & \makecell[c]{0.829 \\ {\scriptsize{(0.808, 0.850)}}} & \makecell[c]{0.829 \\ {\scriptsize{(0.789, 0.866)}}} & \makecell[c]{0.742 \\ {\scriptsize{(0.717, 0.767)}}} \\
    Adults (18+) & \makecell[c]{0.799 \\ {\scriptsize{(0.786, 0.813)}}} & \makecell[c]{0.834 \\ {\scriptsize{(0.814, 0.853)}}} & \makecell[c]{0.915 \\ {\scriptsize{(0.905, 0.924)}}} & \makecell[c]{0.769 \\ {\scriptsize{(0.745, 0.793)}}} \\
    \bottomrule
  \end{tabular}
  \caption[Pediatric cohort results]{\textbf{Predictions for 30-day readmission and selected disease events, by pediatric age group.} \ac{AUROC} with 95\% confidence intervals for each pediatric age cohort and the adult reference cohort for prediction of 30-day readmission, asthma exacerbation, sickle cell crisis, and incident asthma (\autoref{subsubsec:eval-peds}). These are the tabulated values behind \autoref{fig:age_cohort_forest}. Blank cells indicate cohorts excluded from a task because too few patients experience the outcome.}
  \label{tab:full_peds}
\end{table}

\begin{table}[H]
  \centering
  \small
  \setlength{\tabcolsep}{6pt}
  \begin{tabular}{lcccc}
    \toprule
    & \textbf{All} & \textbf{Admissions} & \textbf{Admissions} & \textbf{Admissions} \\
    \textbf{Age cohort} & \textbf{admissions} & \textbf{$<$4 days} & \textbf{$<$7 days} & \textbf{$<$14 days} \\
    \midrule
    Neonates ($<$28 days) & \makecell[c]{2.912 \\ {\scriptsize{(2.685, 3.133)}}} & \makecell[c]{0.601 \\ {\scriptsize{(0.583, 0.624)}}} & \makecell[c]{0.708 \\ {\scriptsize{(0.687, 0.730)}}} & \makecell[c]{0.904 \\ {\scriptsize{(0.871, 0.935)}}} \\
    Infancy (28 days--12 months) & \makecell[c]{3.819 \\ {\scriptsize{(3.481, 4.243)}}} & \makecell[c]{0.945 \\ {\scriptsize{(0.915, 0.979)}}} & \makecell[c]{1.198 \\ {\scriptsize{(1.167, 1.232)}}} & \makecell[c]{1.627 \\ {\scriptsize{(1.586, 1.670)}}} \\
    Toddler (13--23 months) & \makecell[c]{2.181 \\ {\scriptsize{(2.051, 2.316)}}} & \makecell[c]{0.843 \\ {\scriptsize{(0.822, 0.867)}}} & \makecell[c]{1.043 \\ {\scriptsize{(1.021, 1.065)}}} & \makecell[c]{1.323 \\ {\scriptsize{(1.289, 1.357)}}} \\
    Early childhood (2--5 years) & \makecell[c]{2.129 \\ {\scriptsize{(1.969, 2.332)}}} & \makecell[c]{0.817 \\ {\scriptsize{(0.797, 0.838)}}} & \makecell[c]{1.016 \\ {\scriptsize{(0.994, 1.039)}}} & \makecell[c]{1.320 \\ {\scriptsize{(1.287, 1.358)}}} \\
    Middle childhood (6--11 years) & \makecell[c]{2.429 \\ {\scriptsize{(2.243, 2.646)}}} & \makecell[c]{0.917 \\ {\scriptsize{(0.890, 0.949)}}} & \makecell[c]{1.114 \\ {\scriptsize{(1.083, 1.143)}}} & \makecell[c]{1.488 \\ {\scriptsize{(1.406, 1.616)}}} \\
    Early adolescence (12--17 years) & \makecell[c]{2.933 \\ {\scriptsize{(2.666, 3.244)}}} & \makecell[c]{0.975 \\ {\scriptsize{(0.943, 1.006)}}} & \makecell[c]{1.170 \\ {\scriptsize{(1.140, 1.198)}}} & \makecell[c]{1.540 \\ {\scriptsize{(1.501, 1.581)}}} \\
    Adults (18+) & \makecell[c]{2.773 \\ {\scriptsize{(2.599, 2.982)}}} & \makecell[c]{1.082 \\ {\scriptsize{(1.049, 1.114)}}} & \makecell[c]{1.198 \\ {\scriptsize{(1.171, 1.225)}}} & \makecell[c]{1.641 \\ {\scriptsize{(1.600, 1.680)}}} \\
    \bottomrule
  \end{tabular}
  \caption[Pediatric length of stay results]{\textbf{Length-of-stay prediction error for pediatric admissions.} Mean absolute error in days with 95\% confidence intervals for each pediatric age cohort and the adult reference cohort, reported for all admissions and for subsets of admissions <4, <7, and <14 days (\autoref{subsubsec:eval-peds}). These are the tabulated values behind \autoref{fig:age_cohort_los_forest}.}
  \label{tab:full_peds_los}
\end{table}

\begin{table}[H]
  \centering
  \small
  \setlength{\tabcolsep}{4pt}
  \resizebox{\textwidth}{!}{%
  \begin{tabular}{llcccccc}
    \toprule
    & & & \textbf{Numeric} & \textbf{Non-numeric} & \textbf{Medications} & \textbf{Medications} & \\
    \textbf{Age cohort} & \textbf{Encounter} & \textbf{Diagnoses} & \textbf{labs} & \textbf{labs} & \textbf{ordered} & \textbf{reported} & \textbf{Procedures} \\
    \midrule
    \multirow{3}{*}{Neonates ($<$28 days)} & Office visit & \makecell[c]{0.519 \\ {\scriptsize{(0.502, 0.534)}} \\ {\scriptsize BL 0.166}} & \makecell[c]{0.276 \\ {\scriptsize{(0.202, 0.357)}} \\ {\scriptsize BL 0.158}} & \makecell[c]{0.228 \\ {\scriptsize{(0.141, 0.351)}} \\ {\scriptsize BL 0.149}} & \makecell[c]{0.181 \\ {\scriptsize{(0.152, 0.216)}} \\ {\scriptsize BL 0.045}} & \makecell[c]{0.103 \\ {\scriptsize{(0.066, 0.159)}} \\ {\scriptsize BL 0.007}} & \makecell[c]{0.664 \\ {\scriptsize{(0.639, 0.685)}} \\ {\scriptsize BL 0.137}} \\
     & Emergency & \makecell[c]{0.356 \\ {\scriptsize{(0.343, 0.368)}} \\ {\scriptsize BL 0.085}} & \makecell[c]{0.640 \\ {\scriptsize{(0.612, 0.666)}} \\ {\scriptsize BL 0.143}} & \makecell[c]{0.516 \\ {\scriptsize{(0.479, 0.552)}} \\ {\scriptsize BL 0.040}} & \makecell[c]{0.511 \\ {\scriptsize{(0.485, 0.538)}} \\ {\scriptsize BL 0.043}} & \makecell[c]{0.110 \\ {\scriptsize{(0.050, 0.177)}} \\ {\scriptsize BL 0.012}} & \makecell[c]{0.528 \\ {\scriptsize{(0.511, 0.546)}} \\ {\scriptsize BL 0.071}} \\
     & Inpatient & \makecell[c]{0.349 \\ {\scriptsize{(0.336, 0.363)}} \\ {\scriptsize BL 0.163}} & \makecell[c]{0.780 \\ {\scriptsize{(0.765, 0.794)}} \\ {\scriptsize BL 0.289}} & \makecell[c]{0.626 \\ {\scriptsize{(0.603, 0.646)}} \\ {\scriptsize BL 0.099}} & \makecell[c]{0.560 \\ {\scriptsize{(0.544, 0.577)}} \\ {\scriptsize BL 0.090}} & \makecell[c]{0.103 \\ {\scriptsize{(0.051, 0.163)}} \\ {\scriptsize BL 0.061}} & \makecell[c]{0.577 \\ {\scriptsize{(0.565, 0.590)}} \\ {\scriptsize BL 0.120}} \\
    \midrule
    \multirow{3}{*}{Infancy (28 days--12 months)} & Office visit & \makecell[c]{0.609 \\ {\scriptsize{(0.593, 0.624)}} \\ {\scriptsize BL 0.312}} & \makecell[c]{0.407 \\ {\scriptsize{(0.283, 0.537)}} \\ {\scriptsize BL 0.162}} & \makecell[c]{0.228 \\ {\scriptsize{(0.158, 0.309)}} \\ {\scriptsize BL 0.085}} & \makecell[c]{0.258 \\ {\scriptsize{(0.230, 0.285)}} \\ {\scriptsize BL 0.154}} & \makecell[c]{0.166 \\ {\scriptsize{(0.116, 0.226)}} \\ {\scriptsize BL 0.057}} & \makecell[c]{0.811 \\ {\scriptsize{(0.795, 0.825)}} \\ {\scriptsize BL 0.312}} \\
     & Emergency & \makecell[c]{0.444 \\ {\scriptsize{(0.434, 0.453)}} \\ {\scriptsize BL 0.127}} & \makecell[c]{0.542 \\ {\scriptsize{(0.504, 0.574)}} \\ {\scriptsize BL 0.186}} & \makecell[c]{0.467 \\ {\scriptsize{(0.437, 0.496)}} \\ {\scriptsize BL 0.164}} & \makecell[c]{0.457 \\ {\scriptsize{(0.440, 0.473)}} \\ {\scriptsize BL 0.142}} & \makecell[c]{0.160 \\ {\scriptsize{(0.134, 0.188)}} \\ {\scriptsize BL 0.030}} & \makecell[c]{0.495 \\ {\scriptsize{(0.481, 0.508)}} \\ {\scriptsize BL 0.144}} \\
     & Inpatient & \makecell[c]{0.425 \\ {\scriptsize{(0.416, 0.434)}} \\ {\scriptsize BL 0.219}} & \makecell[c]{0.753 \\ {\scriptsize{(0.742, 0.762)}} \\ {\scriptsize BL 0.372}} & \makecell[c]{0.604 \\ {\scriptsize{(0.589, 0.619)}} \\ {\scriptsize BL 0.240}} & \makecell[c]{0.594 \\ {\scriptsize{(0.586, 0.601)}} \\ {\scriptsize BL 0.302}} & \makecell[c]{0.200 \\ {\scriptsize{(0.179, 0.223)}} \\ {\scriptsize BL 0.068}} & \makecell[c]{0.540 \\ {\scriptsize{(0.531, 0.548)}} \\ {\scriptsize BL 0.190}} \\
    \midrule
    \multirow{3}{*}{Toddler (13--23 months)} & Office visit & \makecell[c]{0.540 \\ {\scriptsize{(0.526, 0.554)}} \\ {\scriptsize BL 0.299}} & \makecell[c]{0.453 \\ {\scriptsize{(0.299, 0.579)}} \\ {\scriptsize BL 0.283}} & \makecell[c]{0.280 \\ {\scriptsize{(0.220, 0.348)}} \\ {\scriptsize BL 0.175}} & \makecell[c]{0.242 \\ {\scriptsize{(0.222, 0.264)}} \\ {\scriptsize BL 0.214}} & \makecell[c]{0.155 \\ {\scriptsize{(0.118, 0.190)}} \\ {\scriptsize BL 0.108}} & \makecell[c]{0.760 \\ {\scriptsize{(0.745, 0.775)}} \\ {\scriptsize BL 0.355}} \\
     & Emergency & \makecell[c]{0.425 \\ {\scriptsize{(0.415, 0.434)}} \\ {\scriptsize BL 0.159}} & \makecell[c]{0.492 \\ {\scriptsize{(0.449, 0.534)}} \\ {\scriptsize BL 0.232}} & \makecell[c]{0.453 \\ {\scriptsize{(0.420, 0.486)}} \\ {\scriptsize BL 0.208}} & \makecell[c]{0.466 \\ {\scriptsize{(0.450, 0.482)}} \\ {\scriptsize BL 0.205}} & \makecell[c]{0.130 \\ {\scriptsize{(0.109, 0.153)}} \\ {\scriptsize BL 0.069}} & \makecell[c]{0.509 \\ {\scriptsize{(0.495, 0.523)}} \\ {\scriptsize BL 0.201}} \\
     & Inpatient & \makecell[c]{0.430 \\ {\scriptsize{(0.420, 0.442)}} \\ {\scriptsize BL 0.252}} & \makecell[c]{0.778 \\ {\scriptsize{(0.767, 0.790)}} \\ {\scriptsize BL 0.427}} & \makecell[c]{0.646 \\ {\scriptsize{(0.628, 0.664)}} \\ {\scriptsize BL 0.337}} & \makecell[c]{0.655 \\ {\scriptsize{(0.646, 0.665)}} \\ {\scriptsize BL 0.340}} & \makecell[c]{0.207 \\ {\scriptsize{(0.184, 0.232)}} \\ {\scriptsize BL 0.099}} & \makecell[c]{0.547 \\ {\scriptsize{(0.537, 0.557)}} \\ {\scriptsize BL 0.240}} \\
    \midrule
    \multirow{3}{*}{Early childhood (2--5 years)} & Office visit & \makecell[c]{0.543 \\ {\scriptsize{(0.530, 0.557)}} \\ {\scriptsize BL 0.287}} & \makecell[c]{0.431 \\ {\scriptsize{(0.309, 0.549)}} \\ {\scriptsize BL 0.285}} & \makecell[c]{0.420 \\ {\scriptsize{(0.338, 0.502)}} \\ {\scriptsize BL 0.207}} & \makecell[c]{0.300 \\ {\scriptsize{(0.275, 0.323)}} \\ {\scriptsize BL 0.236}} & \makecell[c]{0.127 \\ {\scriptsize{(0.089, 0.171)}} \\ {\scriptsize BL 0.087}} & \makecell[c]{0.720 \\ {\scriptsize{(0.703, 0.737)}} \\ {\scriptsize BL 0.298}} \\
     & Emergency & \makecell[c]{0.406 \\ {\scriptsize{(0.397, 0.414)}} \\ {\scriptsize BL 0.159}} & \makecell[c]{0.521 \\ {\scriptsize{(0.476, 0.565)}} \\ {\scriptsize BL 0.241}} & \makecell[c]{0.455 \\ {\scriptsize{(0.421, 0.490)}} \\ {\scriptsize BL 0.214}} & \makecell[c]{0.468 \\ {\scriptsize{(0.452, 0.484)}} \\ {\scriptsize BL 0.224}} & \makecell[c]{0.113 \\ {\scriptsize{(0.093, 0.137)}} \\ {\scriptsize BL 0.083}} & \makecell[c]{0.508 \\ {\scriptsize{(0.494, 0.521)}} \\ {\scriptsize BL 0.210}} \\
     & Inpatient & \makecell[c]{0.431 \\ {\scriptsize{(0.422, 0.440)}} \\ {\scriptsize BL 0.261}} & \makecell[c]{0.814 \\ {\scriptsize{(0.807, 0.822)}} \\ {\scriptsize BL 0.455}} & \makecell[c]{0.640 \\ {\scriptsize{(0.626, 0.654)}} \\ {\scriptsize BL 0.337}} & \makecell[c]{0.680 \\ {\scriptsize{(0.674, 0.687)}} \\ {\scriptsize BL 0.351}} & \makecell[c]{0.180 \\ {\scriptsize{(0.160, 0.199)}} \\ {\scriptsize BL 0.111}} & \makecell[c]{0.573 \\ {\scriptsize{(0.565, 0.581)}} \\ {\scriptsize BL 0.250}} \\
    \midrule
    \multirow{3}{*}{Middle childhood (6--11 years)} & Office visit & \makecell[c]{0.567 \\ {\scriptsize{(0.555, 0.580)}} \\ {\scriptsize BL 0.295}} & \makecell[c]{0.501 \\ {\scriptsize{(0.403, 0.592)}} \\ {\scriptsize BL 0.278}} & \makecell[c]{0.448 \\ {\scriptsize{(0.382, 0.518)}} \\ {\scriptsize BL 0.243}} & \makecell[c]{0.326 \\ {\scriptsize{(0.303, 0.348)}} \\ {\scriptsize BL 0.252}} & \makecell[c]{0.116 \\ {\scriptsize{(0.089, 0.146)}} \\ {\scriptsize BL 0.090}} & \makecell[c]{0.711 \\ {\scriptsize{(0.697, 0.725)}} \\ {\scriptsize BL 0.298}} \\
     & Emergency & \makecell[c]{0.371 \\ {\scriptsize{(0.362, 0.380)}} \\ {\scriptsize BL 0.141}} & \makecell[c]{0.559 \\ {\scriptsize{(0.523, 0.594)}} \\ {\scriptsize BL 0.253}} & \makecell[c]{0.492 \\ {\scriptsize{(0.464, 0.519)}} \\ {\scriptsize BL 0.208}} & \makecell[c]{0.480 \\ {\scriptsize{(0.462, 0.498)}} \\ {\scriptsize BL 0.217}} & \makecell[c]{0.078 \\ {\scriptsize{(0.063, 0.098)}} \\ {\scriptsize BL 0.099}} & \makecell[c]{0.512 \\ {\scriptsize{(0.499, 0.525)}} \\ {\scriptsize BL 0.197}} \\
     & Inpatient & \makecell[c]{0.419 \\ {\scriptsize{(0.409, 0.428)}} \\ {\scriptsize BL 0.253}} & \makecell[c]{0.836 \\ {\scriptsize{(0.829, 0.843)}} \\ {\scriptsize BL 0.458}} & \makecell[c]{0.624 \\ {\scriptsize{(0.611, 0.637)}} \\ {\scriptsize BL 0.309}} & \makecell[c]{0.687 \\ {\scriptsize{(0.681, 0.694)}} \\ {\scriptsize BL 0.342}} & \makecell[c]{0.172 \\ {\scriptsize{(0.152, 0.190)}} \\ {\scriptsize BL 0.122}} & \makecell[c]{0.577 \\ {\scriptsize{(0.568, 0.584)}} \\ {\scriptsize BL 0.232}} \\
    \midrule
    \multirow{3}{*}{Early adolescence (12--17 years)} & Office visit & \makecell[c]{0.554 \\ {\scriptsize{(0.540, 0.569)}} \\ {\scriptsize BL 0.290}} & \makecell[c]{0.309 \\ {\scriptsize{(0.231, 0.384)}} \\ {\scriptsize BL 0.218}} & \makecell[c]{0.405 \\ {\scriptsize{(0.341, 0.473)}} \\ {\scriptsize BL 0.229}} & \makecell[c]{0.326 \\ {\scriptsize{(0.297, 0.355)}} \\ {\scriptsize BL 0.235}} & \makecell[c]{0.118 \\ {\scriptsize{(0.087, 0.153)}} \\ {\scriptsize BL 0.109}} & \makecell[c]{0.648 \\ {\scriptsize{(0.627, 0.667)}} \\ {\scriptsize BL 0.272}} \\
     & Emergency & \makecell[c]{0.334 \\ {\scriptsize{(0.324, 0.344)}} \\ {\scriptsize BL 0.127}} & \makecell[c]{0.675 \\ {\scriptsize{(0.652, 0.696)}} \\ {\scriptsize BL 0.319}} & \makecell[c]{0.538 \\ {\scriptsize{(0.515, 0.558)}} \\ {\scriptsize BL 0.216}} & \makecell[c]{0.454 \\ {\scriptsize{(0.437, 0.472)}} \\ {\scriptsize BL 0.167}} & \makecell[c]{0.089 \\ {\scriptsize{(0.065, 0.116)}} \\ {\scriptsize BL 0.100}} & \makecell[c]{0.525 \\ {\scriptsize{(0.513, 0.537)}} \\ {\scriptsize BL 0.188}} \\
     & Inpatient & \makecell[c]{0.366 \\ {\scriptsize{(0.357, 0.375)}} \\ {\scriptsize BL 0.215}} & \makecell[c]{0.857 \\ {\scriptsize{(0.851, 0.863)}} \\ {\scriptsize BL 0.488}} & \makecell[c]{0.622 \\ {\scriptsize{(0.609, 0.634)}} \\ {\scriptsize BL 0.291}} & \makecell[c]{0.645 \\ {\scriptsize{(0.638, 0.652)}} \\ {\scriptsize BL 0.319}} & \makecell[c]{0.158 \\ {\scriptsize{(0.138, 0.180)}} \\ {\scriptsize BL 0.130}} & \makecell[c]{0.588 \\ {\scriptsize{(0.581, 0.595)}} \\ {\scriptsize BL 0.230}} \\
    \midrule
    \multirow{3}{*}{Adults (18+)} & Office visit & \makecell[c]{0.545 \\ {\scriptsize{(0.530, 0.560)}} \\ {\scriptsize BL 0.330}} & \makecell[c]{0.521 \\ {\scriptsize{(0.466, 0.576)}} \\ {\scriptsize BL 0.342}} & \makecell[c]{0.350 \\ {\scriptsize{(0.287, 0.416)}} \\ {\scriptsize BL 0.244}} & \makecell[c]{0.320 \\ {\scriptsize{(0.291, 0.353)}} \\ {\scriptsize BL 0.265}} & \makecell[c]{0.127 \\ {\scriptsize{(0.100, 0.156)}} \\ {\scriptsize BL 0.124}} & \makecell[c]{0.640 \\ {\scriptsize{(0.619, 0.663)}} \\ {\scriptsize BL 0.319}} \\
     & Emergency & \makecell[c]{0.443 \\ {\scriptsize{(0.433, 0.454)}} \\ {\scriptsize BL 0.244}} & \makecell[c]{0.837 \\ {\scriptsize{(0.828, 0.846)}} \\ {\scriptsize BL 0.495}} & \makecell[c]{0.581 \\ {\scriptsize{(0.563, 0.599)}} \\ {\scriptsize BL 0.284}} & \makecell[c]{0.552 \\ {\scriptsize{(0.542, 0.563)}} \\ {\scriptsize BL 0.262}} & \makecell[c]{0.159 \\ {\scriptsize{(0.129, 0.189)}} \\ {\scriptsize BL 0.147}} & \makecell[c]{0.617 \\ {\scriptsize{(0.609, 0.626)}} \\ {\scriptsize BL 0.272}} \\
     & Inpatient & \makecell[c]{0.473 \\ {\scriptsize{(0.466, 0.480)}} \\ {\scriptsize BL 0.301}} & \makecell[c]{0.919 \\ {\scriptsize{(0.916, 0.921)}} \\ {\scriptsize BL 0.617}} & \makecell[c]{0.643 \\ {\scriptsize{(0.633, 0.654)}} \\ {\scriptsize BL 0.353}} & \makecell[c]{0.635 \\ {\scriptsize{(0.630, 0.640)}} \\ {\scriptsize BL 0.367}} & \makecell[c]{0.212 \\ {\scriptsize{(0.194, 0.231)}} \\ {\scriptsize BL 0.178}} & \makecell[c]{0.657 \\ {\scriptsize{(0.652, 0.662)}} \\ {\scriptsize BL 0.326}} \\
    \bottomrule
  \end{tabular}}
  \caption[Pediatric single encounter generation results]{\textbf{Predicted events within single pediatric encounters, by age group.} \ac{PR-AUC} with 95\% confidence intervals by encounter type and data type for each pediatric age cohort and the adult reference cohort (\autoref{subsubsec:eval-peds}). BL indicates the \ac{PR-AUC} of the best lookback-window baseline for that cohort, encounter type and data type. These are the tabulated values behind \autoref{fig:aucprc_dotgrid}.}
  \label{tab:full_peds_encounter_gen}
\end{table}

\begin{table}[H]
  \centering
  \small
  \setlength{\tabcolsep}{6pt}
  \begin{tabular}{lc}
    \toprule
    & \textbf{AUROC} \\
    \textbf{Model} & \textbf{(95\% CI)} \\
    \midrule
    \ugmcurr{} & \makecell[c]{\textbf{0.799} \\ {\scriptsize{(0.786, 0.813)}}} \\
    \ugmprev{} & \makecell[c]{0.772 \\ {\scriptsize{(0.758, 0.786)}}} \\
    \bottomrule
  \end{tabular}
  \caption[30 day readmission results]{\textbf{Prediction of 30-day readmission for adult patients.} AUROC with 95\% confidence intervals for \ugmcurr{} and \ugmprev{} (\autoref{subsubsec:eval-readmission}). Bold marks the higher AUROC.}
  \label{tab:full_readmission}
\end{table}

\begin{table}[H]
  \centering
  \small
  \setlength{\tabcolsep}{6pt}
  \begin{tabular}{lcc}
    \toprule
    & \textbf{\ugmcurr{}} & \textbf{\ugmprev{}} \\
    & \textbf{MAE (days)} & \textbf{MAE (days)} \\
    \textbf{Task} & \textbf{(95\% CI)} & \textbf{(95\% CI)} \\
    \midrule
    Admissions under 7 days  & \makecell[c]{\textbf{1.198} \\ {\scriptsize{(1.171, 1.225)}}} & \makecell[c]{1.238 \\ {\scriptsize{(1.210, 1.266)}}} \\
    Admissions under 14 days & \makecell[c]{\textbf{1.641} \\ {\scriptsize{(1.600, 1.680)}}} & \makecell[c]{1.757 \\ {\scriptsize{(1.718, 1.797)}}} \\
    All admissions           & \makecell[c]{\textbf{2.773} \\ {\scriptsize{(2.599, 2.982)}}} & \makecell[c]{2.850 \\ {\scriptsize{(2.718, 2.989)}}} \\
    \bottomrule
  \end{tabular}
  \caption[Length of stay results]{\textbf{Length-of-stay prediction error for adult admissions.} Mean absolute error in days with 95\% confidence intervals for \ugmcurr{} and \ugmprev{} (\autoref{subsubsec:eval-los}). Bold marks the lower error in each row.}
  \label{tab:full_los}
\end{table}

\begin{table}[H]
  \centering
  \small
  \setlength{\tabcolsep}{6pt}
  \begin{tabular}{lcc}
    \toprule
    & \textbf{\ugmcurr{}} & \textbf{\ugmprev{}} \\
    & \textbf{MAE} & \textbf{MAE} \\
    \textbf{Lab (units)} & \textbf{(95\% CI)} & \textbf{(95\% CI)} \\
    \midrule
    HgbA1c (\%)                  & \makecell[c]{\textbf{0.73} \\ {\scriptsize{(0.71, 0.75)}}} & \makecell[c]{1.07 \\ {\scriptsize{(1.03, 1.10)}}} \\
    HDL cholesterol (mg/dL)      & \makecell[c]{\textbf{5.71} \\ {\scriptsize{(5.50, 5.92)}}} & \makecell[c]{10.61 \\ {\scriptsize{(10.13, 11.08)}}} \\
    LDL cholesterol (mg/dL)      & \makecell[c]{\textbf{19.82} \\ {\scriptsize{(19.20, 20.44)}}} & \makecell[c]{24.26 \\ {\scriptsize{(23.36, 25.21)}}} \\
    Total cholesterol (mg/dL)    & \makecell[c]{\textbf{23.76} \\ {\scriptsize{(23.01, 24.51)}}} & \makecell[c]{33.60 \\ {\scriptsize{(32.23, 34.92)}}} \\
    \bottomrule
  \end{tabular}
  \caption[Lab prediction results]{\textbf{Laboratory value prediction error in adults.} Mean absolute error with 95\% confidence intervals for each predicted lab, reported per model (\autoref{subsubsec:eval-labs}). Bold marks the lower error in each row.}
  \label{tab:full_labs}
\end{table}

\begin{table}[H]
  \centering
  \small
  \setlength{\tabcolsep}{6pt}
  \begin{tabular}{lcc}
    \toprule
    & \textbf{\ugmcurr{}} & \textbf{\ugmprev{}} \\
    & \textbf{AUROC} & \textbf{AUROC} \\
    \textbf{Phenotype (time horizon)} & \textbf{(95\% CI)} & \textbf{(95\% CI)} \\
    \midrule
    COPD (2yr)                        & \makecell[c]{\textbf{0.841} \\ {\scriptsize{(0.822, 0.861)}}} & \makecell[c]{0.839 \\ {\scriptsize{(0.820, 0.858)}}} \\
    CHF (2yr)                         & \makecell[c]{\textbf{0.902} \\ {\scriptsize{(0.884, 0.917)}}} & \makecell[c]{0.891 \\ {\scriptsize{(0.874, 0.906)}}} \\
    Dementia (2yr)                    & \makecell[c]{\textbf{0.926} \\ {\scriptsize{(0.912, 0.939)}}} & \makecell[c]{0.921 \\ {\scriptsize{(0.907, 0.934)}}} \\
    Asthma (2yr)                      & \makecell[c]{0.769 \\ {\scriptsize{(0.745, 0.793)}}} & \makecell[c]{\textbf{0.782} \\ {\scriptsize{(0.759, 0.806)}}} \\
    Alcohol use disorder (2yr)        & \makecell[c]{\textbf{0.807} \\ {\scriptsize{(0.784, 0.829)}}} & \makecell[c]{0.801 \\ {\scriptsize{(0.777, 0.823)}}} \\
    Acute myocardial infarction (2yr) & \makecell[c]{\textbf{0.816} \\ {\scriptsize{(0.794, 0.837)}}} & \makecell[c]{0.801 \\ {\scriptsize{(0.777, 0.823)}}} \\
    \bottomrule
  \end{tabular}
  \caption[Incident disease results]{\textbf{Incident disease prediction in adults.} AUROC with 95\% confidence intervals for each phenotype, reported per model (\autoref{subsubsec:eval-incident}). Bold marks the higher AUROC in each row.}
  \label{tab:full_incident}
\end{table}

\begin{table}[H]
  \centering
  \small
  \setlength{\tabcolsep}{6pt}
  \begin{tabular}{llcc}
    \toprule
    & & \textbf{\ugmcurr{}} & \textbf{\ugmprev{}} \\
    \textbf{Encounter} & \textbf{Data type} & \textbf{PR-AUC (95\% CI)} & \textbf{PR-AUC (95\% CI)} \\
    \midrule
    \multirow{6}{*}{Office visit} & Diagnoses & \makecell[c]{0.545 \\ {\scriptsize{(0.530, 0.560)}}} & \makecell[c]{\textbf{0.548} \\ {\scriptsize{(0.541, 0.556)}}} \\
     & Numeric labs & \makecell[c]{\textbf{0.521} \\ {\scriptsize{(0.466, 0.576)}}} & \makecell[c]{0.421 \\ {\scriptsize{(0.357, 0.491)}}} \\
     & Non-numeric labs & \makecell[c]{0.350 \\ {\scriptsize{(0.287, 0.416)}}} &  \\
     & Medications ordered & \makecell[c]{\textbf{0.320} \\ {\scriptsize{(0.291, 0.353)}}} & \makecell[c]{0.251 \\ {\scriptsize{(0.239, 0.262)}}} \\
     & Medications reported & \makecell[c]{0.127 \\ {\scriptsize{(0.100, 0.156)}}} &  \\
     & Procedures & \makecell[c]{\textbf{0.640} \\ {\scriptsize{(0.619, 0.663)}}} & \makecell[c]{0.638 \\ {\scriptsize{(0.628, 0.649)}}} \\
    \midrule
    \multirow{6}{*}{Emergency} & Diagnoses & \makecell[c]{0.443 \\ {\scriptsize{(0.433, 0.454)}}} & \makecell[c]{\textbf{0.450} \\ {\scriptsize{(0.444, 0.455)}}} \\
     & Numeric labs & \makecell[c]{0.837 \\ {\scriptsize{(0.828, 0.846)}}} & \makecell[c]{\textbf{0.840} \\ {\scriptsize{(0.835, 0.845)}}} \\
     & Non-numeric labs & \makecell[c]{0.581 \\ {\scriptsize{(0.563, 0.599)}}} &  \\
     & Medications ordered & \makecell[c]{\textbf{0.552} \\ {\scriptsize{(0.542, 0.563)}}} & \makecell[c]{0.536 \\ {\scriptsize{(0.530, 0.542)}}} \\
     & Medications reported & \makecell[c]{0.159 \\ {\scriptsize{(0.129, 0.189)}}} &  \\
     & Procedures & \makecell[c]{\textbf{0.617} \\ {\scriptsize{(0.609, 0.626)}}} & \makecell[c]{0.600 \\ {\scriptsize{(0.595, 0.606)}}} \\
    \midrule
    \multirow{6}{*}{Inpatient} & Diagnoses & \makecell[c]{\textbf{0.473} \\ {\scriptsize{(0.466, 0.480)}}} & \makecell[c]{0.469 \\ {\scriptsize{(0.465, 0.474)}}} \\
     & Numeric labs & \makecell[c]{\textbf{0.919} \\ {\scriptsize{(0.916, 0.921)}}} & \makecell[c]{0.899 \\ {\scriptsize{(0.897, 0.901)}}} \\
     & Non-numeric labs & \makecell[c]{0.643 \\ {\scriptsize{(0.633, 0.654)}}} &  \\
     & Medications ordered & \makecell[c]{\textbf{0.635} \\ {\scriptsize{(0.630, 0.640)}}} & \makecell[c]{0.616 \\ {\scriptsize{(0.613, 0.619)}}} \\
     & Medications reported & \makecell[c]{0.212 \\ {\scriptsize{(0.194, 0.231)}}} &  \\
     & Procedures & \makecell[c]{\textbf{0.657} \\ {\scriptsize{(0.652, 0.662)}}} & \makecell[c]{0.619 \\ {\scriptsize{(0.616, 0.623)}}} \\
    \bottomrule
  \end{tabular}
  \caption[Single encounter generation results]{\textbf{Predicted events within single encounters among adult patients.} \ac{PR-AUC} with 95\% confidence intervals by encounter type and data type, reported per model (\autoref{subsubsec:eval-single-encounter}). Non-numeric laboratory tests and reported medications are new data types in \ugmcurr{} and have no \ugmprev{} counterpart. Bold marks the higher \ac{PR-AUC} in each row comparing models.}
  \label{tab:full_encounter_gen}
\end{table}

\end{document}